\documentclass[]{opendatalab}
\usepackage{xcolor}
\usepackage{longtable}
\usepackage{listings}
\usepackage{arydshln}
\usepackage{wrapfig}
\usepackage[numbers]{natbib}
\definecolor{codegreen}{rgb}{0,0.6,0}
\definecolor{codegray}{rgb}{0.5,0.5,0.5}
\definecolor{codepurple}{rgb}{0.58,0,0.82}
\definecolor{backcolour}{rgb}{0.95,0.95,0.92}
\definecolor{promptcolor}{HTML}{D1D0F2}
\definecolor{promptcolorheader}{HTML}{bdbcec}
\newcommand{\promptbox}[2]{
\begin{tcolorbox}[
top=0.3em,bottom=0.3em,left=0.5em,right=0.5em,
toptitle=0.3em,bottomtitle=0.2em,boxsep=0pt,
colframe=promptcolorheader,colback=promptcolor!50,boxrule=0.5pt,
]
\footnotesize
\end{tcolorbox}
}
\lstdefinestyle{mystyle}{
    backgroundcolor=\color{backcolour},   
    commentstyle=\color{codegreen},
    keywordstyle=\color{magenta},
    numberstyle=\tiny\color{codegray},
    stringstyle=\color{codepurple},
    basicstyle=\ttfamily\footnotesize,
    breakatwhitespace=false,         
    breaklines=true,                 
    captionpos=b,                    
    keepspaces=true,                 
    numbers=left,                    
    numbersep=5pt,                  
    showspaces=false,
    showstringspaces=false,
    showtabs=false,                  
    tabsize=2
}

\usepackage{xspace}
\newcommand{\method }{Middo\xspace}
\definecolor{bluex}{rgb}{0.27, 0.42, 0.81}
\definecolor{purplex}{HTML}{9564bf}
\definecolor{red3}{HTML}{C52A20}
\definecolor{red2}{HTML}{B36A6F}
\definecolor{red1}{HTML}{FFb5b5}
\definecolor{purple}{HTML}{B36A6F}

\title{Middo: Model-Informed Dynamic Data Optimization for Enhanced LLM Fine-Tuning via Closed-Loop Learning}

\author[1]{Zinan Tang}
\author[1]{Xin Gao}
\author[1]{Qizhi Pei}
\author[1]{Zhuoshi Pan}
\author[1]{Mengzhang Cai}
\author[1]{Jiang Wu}
\author[1*]{Conghui He}
\author[1*]{Lijun Wu}

\affiliation[1]{OpenDataLab, Shanghai Artificial Intelligence Laboratory}

\abstract{
Supervised Fine-Tuning (SFT) Large Language Models (LLMs) fundamentally rely on high-quality training data.
While data selection and data synthesis are two common strategies to improve data quality, existing approaches often face limitations in static dataset curation that fail to adapt to evolving model capabilities.
In this paper, we introduce \textbf{\method}, a self-evolving \textbf{M}odel-\textbf{i}nformed \textbf{d}ynamic \textbf{d}ata \textbf{o}ptimization framework that uses model-aware data selection and context-preserving data refinement. 
Unlike conventional one-off filtering/synthesis methods, our framework establishes a closed-loop optimization system:
(1) A self-referential diagnostic module proactively identifies suboptimal samples through tri-axial model signals-\textit{loss patterns (complexity)}, \textit{embedding cluster dynamics (diversity)}, and \textit{self-alignment scores (quality)};
(2) An adaptive optimization engine then transforms suboptimal samples into pedagogically valuable training points while preserving semantic integrity;
(3) This optimization process continuously evolves with the model's capability through dynamic learning principles.
Experiments on multiple benchmarks demonstrate that our \method consistently enhances the quality of seed data and boosts LLMs' performance, improving accuracy by $7.15\%$ on average while maintaining the original dataset scale.
This work establishes a new paradigm for sustainable LLM training through dynamic human-AI co-evolution of data and models.
}

\date{\today}
\correspondence{Conghui He, \email{heconghui@pjlab.org.cn}; Lijun Wu, \email{lijunwu@pjlab.org.cn}}

\metadata[Code]{\url{https://github.com/Word2VecT/Middo}}
\metadata[Data \& Models]{\url{https://huggingface.co/collections/Word2Li}}

\begin{document}

\maketitle

\section{Introduction}

Large Language Models (LLMs) have revolutionized artificial intelligence by achieving state-of-the-art performance across diverse domains, from natural language understanding~\citep{zhou2023instructionfollowingevaluationlargelanguage,hendrycks2021measuring} to mathematical reasoning~\citep{cobbe2021trainingverifierssolvemath,hendrycks2021measuring2} and code generation~\citep{chen2021evaluatinglargelanguagemodels,austin2021programsynthesislargelanguage}.
This success is largely attributed to Supervised Fine-Tuning (SFT), where models undergo rigorous training on high-quality, human-aligned datasets to ensure outputs closely match human expectations.
Crucially, the quality of these datasets directly dictates the model’s ultimate capabilities: noisy or suboptimal training data can lead to degraded performance, while meticulously curated data unlocks advanced reasoning, generalization, and robustness.
As LLMs scale, the adage ``garbage in, garbage out'' becomes increasingly important—highlighting the urgent need for systematic methods to optimize training data quality.

Existing approaches primarily fall into two categories to improve data quality: data selection~\citep{cao2024instruction, zhou2023datasetquantization, li2024selfalignment, jia2024boostingllmlearningdata, zhou2024davirdataselectionimplicit,li2024oneshotlearninginstructiondata, li-etal-2024-selective} and data synthesis~\citep{dai2023auggptleveragingchatgpttext, wang2023lets, mukherjee2023orcaprogressivelearningcomplex, xu2025magpie, liu2024best, gao-etal-2025-strategic}. 
Data selection methods filter raw datasets using heuristic rules (e.g., length filters)~\citep{10.5555/3692070.3694581} or statistical metrics like perplexity (PPL)~\citep{liu2024best} and Instruction-Following Difficulty (IFD)~\citep{li-etal-2024-quantity} to retain ``high-quality'' samples.
Conversely, data synthesis leverages advanced LLMs (e.g., GPT-4~\citep{achiam2023gpt}) to generate new training examples, often through prompting or distillation~\citep{ li2024learningcommitteereasoningdistillation}.
While both strategies improve data quality, they suffer from critical limitations.
Selection methods are typically static, applying fixed criteria that ignore the evolving needs of the model during training.
Similarly, synthesis approaches often discard original data, wasting potentially valuable information, and risk generating distributionally narrow or redundant examples. 
These one-time data curation methods fail to adaptively refine data along with the model’s progress.

\begin{wrapfigure}{r}{0.5\textwidth}
    \vspace{-10pt}
    \centering
    \includegraphics[width=\linewidth]{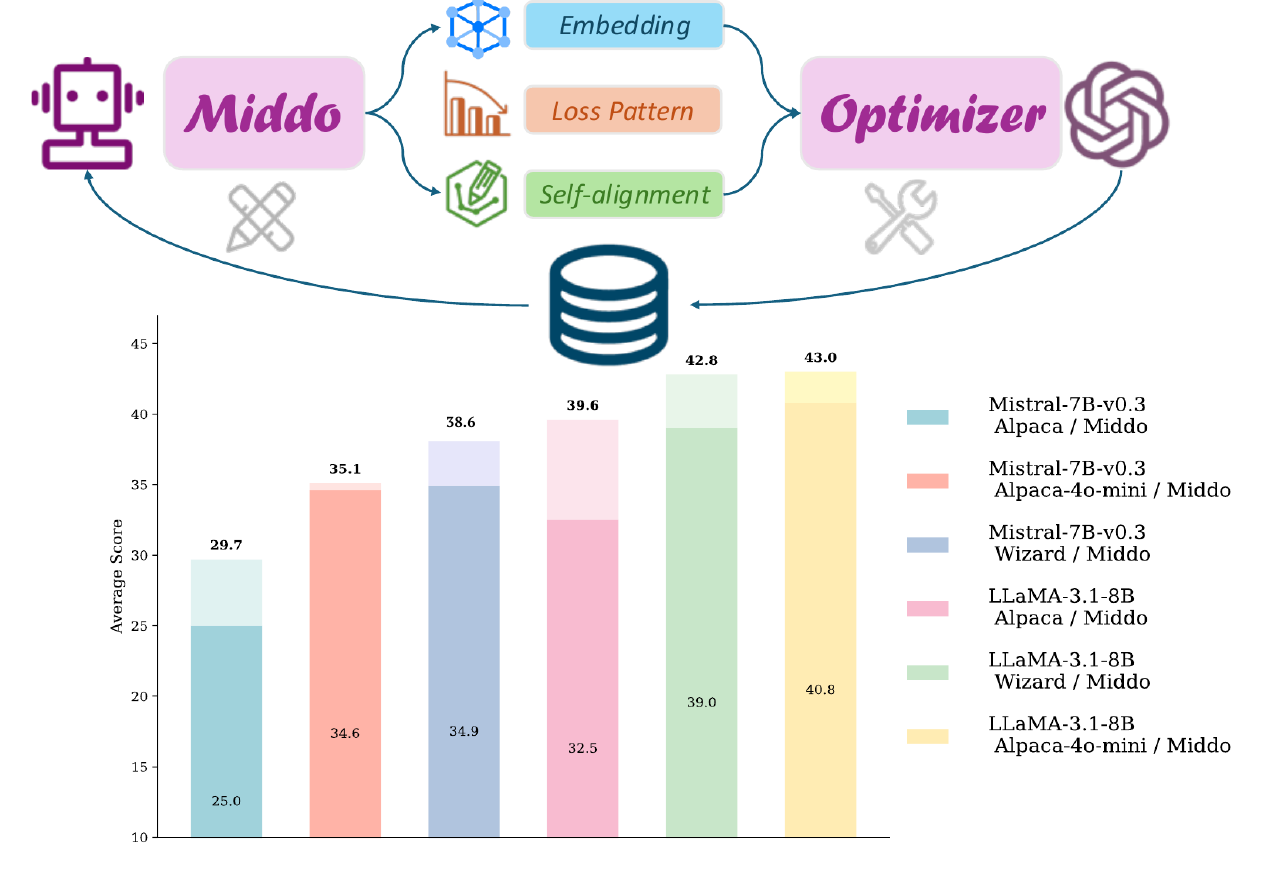}
    \caption{Comparison of different dataset and different models before and after \method optimization.}
    \vspace{-5pt}
\end{wrapfigure}

To overcome these limitations, we propose \textbf{\method }, Model-informed Dynamic Data Optimization, a self-evolving framework that unifies model-aware data selection with context-preserving data refinement.
Unlike static approaches, \method establishes a closed-loop optimization system where data curation dynamically adapts to the model’s evolving capabilities.
The framework operates through three core mechanisms:
(1) A self-referential diagnostic module that proactively identifies suboptimal training samples using three model signals: \textit{loss patterns} (to detect \textbf{complexity} mismatches between data and model proficiency), \textit{embedding cluster dynamics} (to assess \textbf{diversity} gaps in the latent space), and \textit{self-alignment scores} (to evaluate data \textbf{quality} against the model’s own knowledge).
(2) An adaptive optimization engine that transforms these suboptimal samples into pedagogically valuable training points.
For example, overly complex samples may be simplified through stepwise decomposition, while low-diversity clusters are enriched with controlled extension—all while preserving the original data’s semantic intent.
(3) A dynamic principle that iteratively updates the training dataset based on the model’s progress, ensuring that data difficulty and diversity scale with the model’s capabilities.
By integrating these components, \method not only maximizes the utility of existing data but also bridges the gap between static data curation and adaptive model training.

Experiments across multiple benchmarks demonstrate \method ’s effectiveness especially on low-quality datasets. Models trained with \method optimized data achieve consistent performance gains over baselines, improving accuracy by $7.15\%$ on average while maintaining the original dataset scale.
Notably, \method-trained models exhibit stronger abilities to address hard problems, solving more than three times the number of challenging test problems (e.g., MATH, GPQA) compared to models trained on static datasets.
These results validate that sustainable LLM advancement requires co-evolving data and models—a paradigm shift from today’s disjointed curation practices.

\section{Related Work}

\subsection{Synthetic Data Generation}

Synthetic data generation is a key technique for augmenting LLM fine-tuning.
Early methods~\citep{edunov-etal-2018-understanding, wieting-gimpel-2018-paranmt} introduce perturbation-based approaches to enhance data diversity, using character-level~\citep{belinkov2018synthetic} and word-level~\citep{wei-zou-2019-eda} modifications.
These methods rely on fixed transformation rules, limiting adaptability.

LLMs have been leveraged for scalable data synthesis~\citep{sudalairaj2024lablargescalealignmentchatbots,jung-etal-2024-impossible,dai2023auggptleveragingchatgpttext, wang2023lets, mukherjee2023orcaprogressivelearningcomplex, xu2025magpie, liu2024best,li2025synthetic}.
Self-instruct methods~\citep{wang-etal-2023-self-instruct} generate instruction-response pairs, while Evol-Instruct~\citep{xu2024wizardlm} and Auto-Evol-Instruct~\citep{zeng-etal-2024-automatic} refine data complexity iteratively.
However, these methods remain static, failing to adapt as models improve.
Recent approaches integrate model feedback into data generation~\citep{anonymous2025dataenvgym, cao2025condorenhancellmalignment, anonymous2025isheep, li2024learningcommitteereasoningdistillation}, incorporating student model signals for adaptive synthesis.
LLM2LLM~\citep{lee-etal-2024-llm2llm} is an iterative data augmentation strategy that enhances low-data fine-tuning by using a teacher LLM to generate synthetic training data from incorrect student LLM predictions and I-SHEEP~\citep{park2024llamaduollmopspipelineseamless} uses an iterative self-enhancement paradigm.

\subsection{Data Selection}

Data selection is crucial for LLM fine-tuning, as high-quality and informative data directly impacts model performance~\citep{zhou2023lima, xu2023rethinkinginstructionqualitylift}.
Early heuristic-based methods rely on surface-level statistics like item frequency~\citep{10.5555/3455716.3455856} and repetition count~\citep{laurenon2022the}, but they also lack adaptability to model evolution.

Recent work explores LLM-driven data selection, optimizing for quality, diversity, and complexity~\citep{cao2024instruction, zhou2023datasetquantization, li2024selfalignment, jia2024boostingllmlearningdata, zhou2024davirdataselectionimplicit,li2024oneshotlearninginstructiondata, li-etal-2024-selective,du2023modsmodelorienteddataselection,kung2023active}.
The IFD metric~\citep{li-etal-2024-quantity} enables models to self-select training instances by comparing loss with and without the instruction, while other methods~\citep{yu2024wavecoderwidespreadversatileenhancement,colombo2024saullm7bpioneeringlargelanguage,lu2024instag,zhao-etal-2024-tree} use LLM self-assessment for efficiency. Further advancements integrate LLM-based evaluation mechanisms.
AlpaGasus~\citep{chen2024alpagasus} and LIFT~\citep{xu2023rethinkinginstructionqualitylift} use structured prompts for data assessment, while DEITA~\citep{liu2024what} introduces a multi-dimensional scoring system based on complexity and quality.

\section{Methodology}

\begin{figure}[!htb]
  \centering 
  \includegraphics[width=0.9\linewidth]{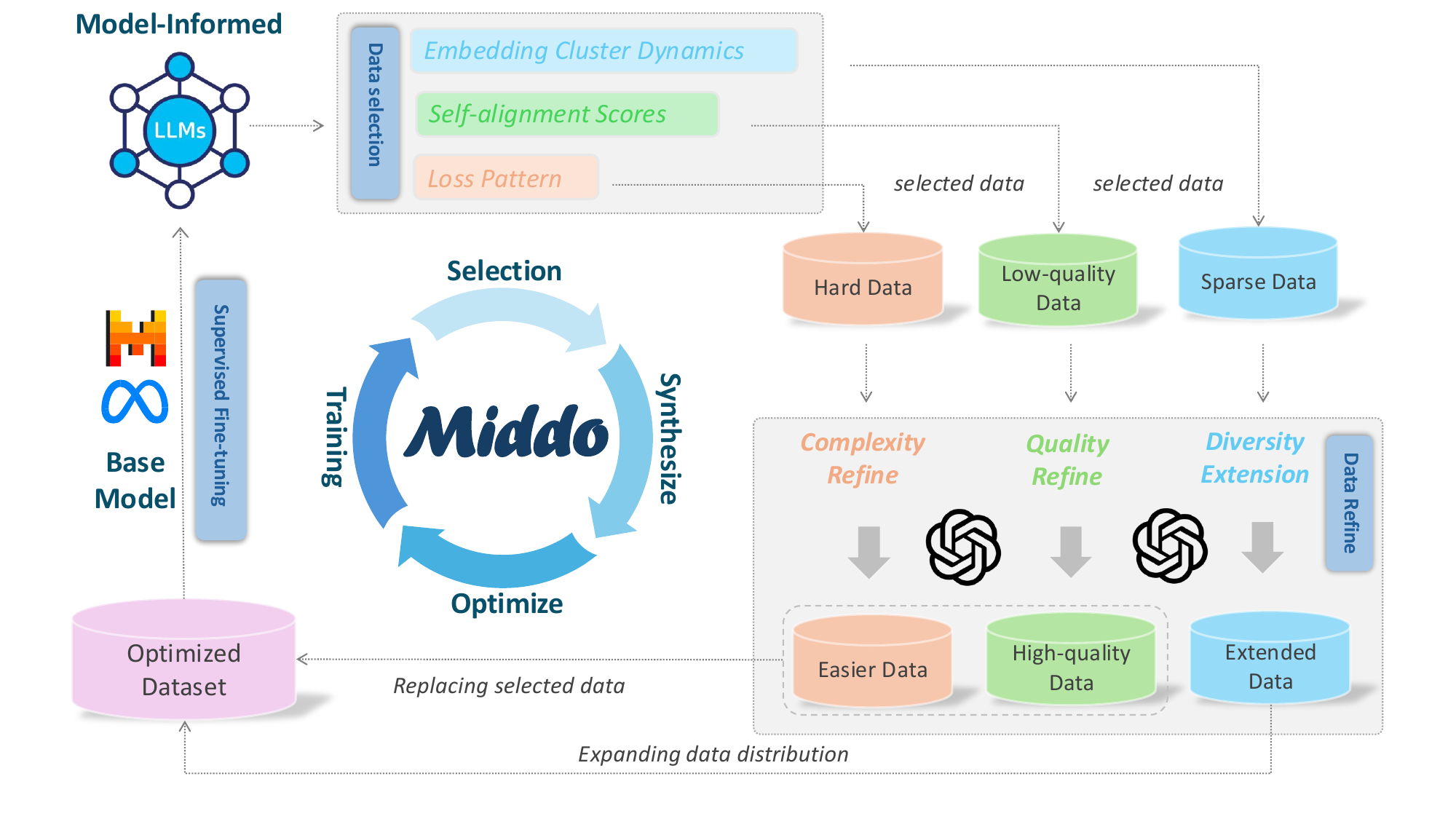}
  \caption{The \method pipeline: a closed-loop, iterative dynamic optimization framework for LLM fine-tuning. It comprises three core modules that leverage model feedback: \textit{Loss Patterns} identify overly complex samples, which are then simplified; \textit{Self-alignment Scores} evaluate data quality, transforming low-quality samples into high-quality ones; and \textit{Embedding Cluster Dynamics} detect sparse data points and expand the data distribution through targeted augmentation. \method ensure the training set continually evolves to better align with the model’s capabilities.}
  \label{fig:pipeline}
\end{figure}

An overview of our \method is shown in Figure~\ref{fig:pipeline}.
We first introduce the overall pipeline of \method in Section~\ref{sec:31_pipeline}, then elaborate on the three core components: complexity optimization (Section~\ref{sec:32_co}), diversity optimization (Section~\ref{sec:33_do}), and quality optimization (Section~\ref{sec:34_qo}).

\subsection{\method Pipeline}
\label{sec:31_pipeline}
As depicted in Figure~\ref{fig:pipeline}, our \method framework establishes an iterative data-model co-evolution loop driven by tri-axial signal analysis, along with three interconnected data optimization mechanisms, each targeting distinct dimensions of training sample selection:
(1) \textit{Loss patterns}, to identify samples with mismatched complexity (overly challenging) relative to the current model’s capability through loss trajectory analysis.
(2) \textit{Embedding cluster dynamics}, to detect coverage gaps in the semantic space, ensuring balanced conceptual representation.
(3) \textit{Self-alignment scores}, for quality filtering to leverage the model’s self-evaluation capacity to flag low-confidence or inconsistent responses through automated alignment scoring.

At each iteration, these parallel signal analyzers jointly select suboptimal samples, which are then regenerated through context-aware synthesis—preserving original semantic intent while enhancing pedagogical value.
The refined dataset immediately feeds back into model training, creating a dynamic feedback loop where improved model capabilities inform subsequent optimization cycles.
Notably, the optimized dataset remains similar in data size, without extending large data synthesis, leading to an efficient data optimization.
This self-referential mechanism ensures continuous alignment between data characteristics and model evolution.
The following sections systematically elaborate on the implementation of each signal-specific optimization module and their synergistic integration.

\subsection{\textit{Loss Patterns}: Complexity Optimization}
\label{sec:32_co}
\paragraph{Complexity Selection.}
Complexity reflects the ``difficulty'' or ``compositionality'' of data.
A good dataset usually requires a smooth complexity distribution of data for training~\citep{havrilla2024surveying, shen2023efficient}. Therefore, we introduce \textit{Loss Patterns}, which targets overly challenging samples by modifying them to maintain a balanced and learnable training set~\citep{zhao-etal-2024-tree}.
During fine-tuning, the loss for a sample $(X_i, Y_i)$ is computed as the likelihood of predicting successive tokens given the instruction $X_i$ and its context.
We denote the loss before and after training as $\mathcal{L}_{\text{pre}}(X_i, Y_i)$ and $\mathcal{L}_{\text{post}}(X_i, Y_i)$, respectively.

Intuitively, we consider both the loss before and after training to select the complex data.
Specifically, we classify samples based on their loss evolution: samples with both low $\mathcal{L}_{\text{pre}}$ and $\mathcal{L}_{\text{post}}$ are considered easy, while those with high values in both metrics remain difficult, indicating excessive complexity. 
A sample is included in the complex subset $\mathcal{D}^{\text{hard}}$ if its $\mathcal{L}_{\text{pre}}$ and $\mathcal{L}_{\text{post}}$ both exceed the thresholds $\tau_{\text{pre}}$ and $\tau_{\text{post}}$, respectively.
For adaptive refinement, the thresholds are dynamically computed.
See Appendix~\ref{apx:thr} for details on the dynamic threshold settings used throughout the paper.

\paragraph{Complexity Optimization.}
For complex data optimization, instead of discarding difficult samples, we transform them into simpler, more manageable forms.
Specifically, we replace samples in $\mathcal{D}^{\text{hard}}$ with their simplified counterparts, $\mathcal{D}^{\text{hard}^\prime}$.
This is achieved by an automatic process in which a LLM analyzes and summarizes the complex instructions~\citep{zeng-etal-2024-automatic}, then simplifies them step by step while preserving the core educational content.
An example is shown in Appendix Figure~\ref{fig:complexity}.
This iterative transformation process updates the dataset by replacing overly complex samples with refined versions that offer more effective training samples.
As training continues, this adaptive approach ensures a continuous alignment between data complexity and model capability.

\subsection{\textit{Embedding Cluster Dynamics}: Diversity Optimization}
\label{sec:33_do}
\paragraph{Diversity Selection.}
Diversity is crucial for ensuring broad concept coverage and a uniform data distribution.
\textit{Embedding Cluster Dynamics} identifies sparse data points that signal underrepresented regions in the dataset.
We extract sentence embeddings from the last hidden layer $\mathcal{H}^{(L)}$ of the model trained in the previous iteration, using average pooling, then compute the cosine similarity between each data point and find the $k$-nearest neighbors $\mathcal{N}_k(X_i)$ for each data $X_i$.
A lower average cosine similarity among these neighbors $\mathcal{N}_k(X_i)$ indicates the data is positioned in a sparser region. 
Thus, the data points whose average cosine similarity score (diversity score) is below a threshold are selected for optimization.

\paragraph{Diversity Optimization.}
To enhance diversity-balanced distribution, we augment the sparse subset $\mathcal{D}^{\text{sparse}}$ by incorporating examples from their corresponding $\mathcal{N}_k(X_i)$ as demonstrations to generate new samples.
This process generates an expanded set $\mathcal{D}^{\text{sparse}{\prime}}$, which is then integrated back into the dataset.
An instance can be found in Appendix Figure~\ref{fig:diversity}. 
This structured augmentation strategy ensures that the data distribution becomes both broader and more balanced, ultimately improving the model’s generalization.

\subsection{\textit{Self-alignment Scores}: Quality Optimization}
\label{sec:34_qo}
\paragraph{Quality Selection.}
High-quality data is essential for fine-tuning, as poor-quality samples can degrade performance~\citep{zhou2023lima}.
To reduce manual annotation costs, many approaches use the LLM-as-a-Judge paradigm~\citep{chen2024alpagasus, xu2023rethinkinginstructionqualitylift}.
To achieve this, instead of relying on an external judge, we leverage the fine-tuned model itself to assess data quality via \textit{Self-alignment Scores}, effectively incorporating the model's own feedback.
Specifically, for each instruction-response pair $(X_i, Y_i)$ in $\mathcal{D}$, the model generate scores $\mathcal{S}{\pi}^{\text{ins.}}(X_i)$ for instruction and $\mathcal{S}{\pi}^{\text{res.}}(X_i, Y_i)$ for instruction-response pair based on three key metrics $\pi$ from AlignBench~\citep{liu-etal-2024-alignbench}: \textit{Clarity}, \textit{Completeness}, and \textit{Factuality}.
The final quality score $\mathcal{S}(X_i, Y_i)$ is obtained by averaging these scores. 
These samples with scores below a similar dynamic threshold are identified as low-quality, forming the seed dataset $\mathcal{D}^{\text{low}}$.

\paragraph{Quality Optimization.}
To refine $\mathcal{D}^{\text{low}}$, we use LLMs to automatically analyze and improve these samples via tailored evolution strategies (prompt templates and examples are provided in the Appendix Figure~\ref{fig:quality}).
This process converts low-quality samples into higher-quality versions, denoted as $\mathcal{D}^{\text{low}\prime}$.
The dataset is then updated by replacing the original low-quality samples with $\mathcal{D}^{\text{low}\prime}$, maintaining the dataset size while progressively enhancing its overall quality.

In each iteration, after the three data selection and optimization processes described above, the optimized dataset is then fed back for the next round of model training.

\section{Experiment}

\subsection{Settings}

\paragraph{Data Optimization Configurations.}
We conduct optimization on the Alpaca~\citep{alpaca} and WizardLM~\citep{xu2024wizardlm} datasets.
For a fair comparison, we also include a rewritten version of Alpaca, where responses are generated by GPT-4o-mini, in our optimization process.
Each dataset undergoes three iterations of optimization.
Demonstrating that our method does not require a powerful external model, we synthesize data using DataDreamer~\citep{patel2024datadreamer} with GPT-4o-mini, setting both \texttt{temperature} and \texttt{top\_p} to 1.0 to ensure diversity.
A detailed analysis of the computational cost is provided in Appendix~\ref{sec:computational_cost}, and the effects of the number of neighbors and iteration counts are discussed in Appendix~\ref{sec:hyper_para_analy}.

\paragraph{Training and Evaluation Settings.}
We fine-tune LLaMA-3.1-8B~\citep{grattafiori2024llama3herdmodels} and Mistral-7B-v0.3~\citep{jiang2023mistral7b} using LLaMA-Factory~\citep{zheng2024llamafactory} with the specific hyperparameters detailed in Appendix~\ref{sec:hyper}.
For each iteration of \method's optimization, the base model is fine-tuned for one epoch on the dataset optimized in that specific iteration to mitigate the risk of overfitting to the data~\citep{lee-etal-2024-llm2llm, anonymous2025isheep}.
Evaluation is conducted using OpenCompass~\citep{2023opencompass}, with vLLM~\citep{10.1145/3600006.3613165} for acceleration.
To validate the effectiveness and generalization capabilities of our approach, we assess model capabilities in general knowledge using IFEval~\citep{zhou2023instructionfollowingevaluationlargelanguage} and MMLU~\citep{hendrycks2021measuring}; mathematical problem-solving on GSM8K~\citep{cobbe2021trainingverifierssolvemath} and MATH~\citep{hendrycks2021measuring2}; code generation on HumanEval~\citep{chen2021evaluatinglargelanguagemodels} and MBPP~\citep{austin2021programsynthesislargelanguage}; and commonsense reasoning on Hellaswag~\citep{zellers-etal-2019-hellaswag} and GPQA~\citep{rein2024gpqa}.

\subsection{Main Results}

The evaluation results on all benchmarks over various data iterations and models are presented in Table~\ref{tab:main_results}. We can see that \method consistently enhances model performance across all benchmarks, achieving an average accuracy increase of $7.15\%$ over three iterations on the Alpaca dataset based on LLaMA-3.1-8B, all while preserving the original data scale.
Moreover, when extending our experiments to Mistral-7B-v0.3, we observed an average improvement of $4.75\%$, further underscoring the robustness and adaptability of our framework across different model architectures.

\begin{table}[!htb]
    \centering
    \resizebox{\textwidth}{!}{
        \begin{tabular}{@{}lcccccccccc@{}}
        \toprule
        \multicolumn{2}{c}{\multirow{2}{*}{\textbf{Setting}}} & \multicolumn{2}{c}{\textbf{General}} & \multicolumn{2}{c}{\textbf{Math}} & \multicolumn{2}{c}{\textbf{Code}} & \multicolumn{2}{c}{\textbf{Reasoning}} & \multirow{2}{*}{\textbf{Average}} \\ 
        \cmidrule(r){3-4} \cmidrule(r){5-6} \cmidrule(r){7-8} \cmidrule(r){9-10}
         &  & MMLU & IFEval & GSM8K & MATH & HumanEval & MBPP & Hellaswag & GPQA & \\ 
        \midrule
        \multicolumn{11}{c}{\textit{Base Model: LLaMA-3.1-8B}} \\
        \midrule
        \noalign{\vskip 0.5ex}
        \multirow{4}{*}{\textbf{Alpaca}} 
            & \textit{init}  & 47.46 & 41.09 & 35.63 & 4.96  & 39.63 & 37.40 & 48.11 & 5.56  & 32.48 \\ 
        \noalign{\vskip 0.5ex}\cdashline{2-11}\noalign{\vskip 0.5ex}
            & \textit{iter1} & 50.13 & 45.77 & 43.67 & 10.62 & 40.24 & 39.20 & 56.37 & 13.64 & 37.45 \\
            & \textit{iter2} & 41.82 & 44.63 & 50.11 & 12.40 & 39.63 & 41.40 & 59.22 & 18.18 & \underline{38.42} \\
            & \textit{iter3} & 51.32 & 43.20 & 51.18 & 12.92 & 39.63 & 41.80 & 58.78 & 16.67 & \textbf{39.63} \\
        \midrule
        \multirow{4}{*}{\shortstack{\textbf{Alpaca} \\ \textbf{4o-mini}}} 
            & \textit{init}  & 32.82 & 44.04 & 57.09 & 17.78 & 51.22 & 45.20 & 53.70 & 24.24 & 40.76 \\ 
        \noalign{\vskip 0.5ex}\cdashline{2-11}\noalign{\vskip 0.5ex}
            & \textit{iter1} & 41.09 & 43.47 & 54.21 & 17.34 & 51.22 & 46.00 & 59.11 & 21.72 & 41.77 \\
            & \textit{iter2} & 44.69 & 47.96 & 57.62 & 18.50 & 52.44 & 45.40 & 57.37 & 19.70 & \textbf{42.96} \\
            & \textit{iter3} & 38.58 & 48.11 & 58.68 & 18.30 & 46.95 & 46.80 & 52.37 & 28.79 & \underline{42.32} \\
        \midrule
        \multirow{4}{*}{\textbf{Wizard}} 
            & \textit{init}  & 46.12 & 46.14 & 53.30 & 12.72 & 40.24 & 48.00 & 53.05 & 12.12 & 38.96 \\ 
        \noalign{\vskip 0.5ex}\cdashline{2-11}\noalign{\vskip 0.5ex}
            & \textit{iter1} & 48.39 & 50.11 & 54.44 & 13.80 & 46.95 & 45.00 & 63.54 & 20.20 & \textbf{42.80} \\
            & \textit{iter2} & 48.86 & 49.48 & 55.12 & 13.90 & 48.78 & 45.20 & 58.63 & 18.18 & \underline{42.29} \\
            & \textit{iter3} & 47.18 & 50.79 & 54.51 & 11.70 & 43.29 & 45.40 & 62.97 & 20.20 & 42.01 \\
        \specialrule{1.5pt}{0pt}{0pt}
        \multicolumn{11}{c}{\textit{Base Model: Mistral-7B-v0.3}} \\ 
        \midrule
        \noalign{\vskip 0.5ex}
        \multirow{4}{*}{\textbf{Alpaca}} 
            & \textit{init}  & 27.66 & 43.22 & 22.21 & 3.88 & 29.27 & 28.80 & 44.17 & 0.51  & 24.97 \\ 
        \noalign{\vskip 0.5ex}\cdashline{2-11}\noalign{\vskip 0.5ex}
            & \textit{iter1} & 31.31 & 45.62 & 29.57 & 5.82 & 30.49 & 33.80 & 42.73 & 14.65 & 29.25 \\
            & \textit{iter2} & 26.87 & 49.46 & 31.69 & 6.84 & 31.71 & 31.00 & 53.95 & 5.56  & \underline{29.64} \\
            & \textit{iter3} & 38.73 & 44.01 & 34.80 & 6.64 & 26.22 & 31.40 & 44.86 & 11.11 & \textbf{29.72} \\
        \midrule
        \multirow{4}{*}{\shortstack{\textbf{Alpaca} \\ \textbf{4o-mini}}} 
            & \textit{init}  & 31.56 & 43.14 & 44.88 & 9.64  & 42.07 & 37.80 & 46.25 & 21.21 & \underline{34.56} \\ 
        \noalign{\vskip 0.5ex}\cdashline{2-11}\noalign{\vskip 0.5ex}
            & \textit{iter1} & 31.33 & 47.93 & 45.19 & 8.72  & 37.20 & 41.32 & 41.32 & 19.70 & 34.09 \\
            & \textit{iter2} & 28.83 & 47.92 & 48.90 & 11.34 & 35.37 & 38.40 & 42.63 & 27.27 & \textbf{35.08} \\
            & \textit{iter3} & 28.96 & 50.78 & 48.60 & 10.10 & 32.32 & 39.00 & 32.95 & 20.20 & 32.86 \\
        \midrule
        \multirow{4}{*}{\textbf{Wizard}} 
            & \textit{init}  & 40.71 & 50.95 & 44.96 & 8.10  & 35.98 & 35.60 & 53.98 & 9.09  & 34.92 \\ 
        \noalign{\vskip 0.5ex}\cdashline{2-11}\noalign{\vskip 0.5ex}
            & \textit{iter1} & 41.39 & 51.18 & 44.43 & 9.44  & 37.80 & 38.60 & 59.01 & 17.17 & 37.38 \\
            & \textit{iter2} & 33.87 & 51.71 & 47.08 & 9.26 & 39.02 & 38.40 & 66.18 & 19.70 & \underline{38.15} \\
            & \textit{iter3} & 43.26 & 49.80 & 41.09 & 10.02 & 41.46 & 34.60 & 66.02 & 22.22 & \textbf{38.56} \\
        \bottomrule
        \end{tabular}
    }
    \caption{
    Performance comparison on different benchmarks using \textit{LLaMA-3.1-8B} and \textit{Mistral-7B-v0.3} as base models.
    We use Alpaca, Alpaca-4o-mini, and Wizard as the optimization datasets for \method.
    The \textit{init} means training on the original dataset, while \textit{iter} means training on the optimized dataset. 
    Both \textit{init} and \textit{iter} settings are trained for one epoch.
    The best performance on average is highlighted in \textbf{bold} and the second best is \underline{underlined}.}
    \label{tab:main_results}
\end{table}

On the Alpaca dataset, the average score increased progressively with each iteration.
Across the MMLU, GSM8K, MATH, and MBPP benchmarks, we observed consistent, step-by-step improvements over multiple iterations. This showcases the versatility of our approach, which excels in general capabilities, mathematics, and coding.
Notably, accuracy on GSM8K improved by $15.55\%$, and Hellaswag saw an $11.11\%$ increase when evaluated on the LLaMA-3.1-8B model.
For Mistral-7B-v0.3, we observed an $11.07\%$ improvement on MMLU, a $12.59\%$ increase on GSM8K, and a $10.6\%$ gain on GPQA.
These results underscore the effectiveness of our method in driving performance gains and highlight the cumulative benefit of our iterative optimization process.

\paragraph{Further Validation on 4o-mini Rewritten Data.}
Steady improvements observed on the 4o-mini rewritten Alpaca dataset—averaging a $2.2\%$ increase overall, with MMLU showing an impressive $11.87\%$ boost—demonstrate that these gains are not merely a result of using 4o-mini data.
This illustrates that our framework intrinsically enhances data quality and model performance.
Importantly, we achieve these improvements without resorting to stronger variants such as GPT-4o~\citep{hurst2024gpt}, reinforcing the robustness and general applicability of our method.

\paragraph{Initial Dataset Quality.}
Our experiments reveal that higher-quality datasets require fewer modifications to reach optimal performance.
On LLaMA-3.1-8B, for instance, while the Alpaca dataset achieves peak performance at third iteration, the 4o-mini rewritten Alpaca required only two iterations, and the Wizard dataset reaches its best performance in just one round.

\paragraph{Comparison with Other Works.}

\begin{table*}[htbp]
    \centering
    \resizebox{\textwidth}{!}{
        \begin{tabular}{ccccccccccc}
        \toprule
        \multirow{2}{*}{\textbf{Method}} & \multirow{2}{*}{\textbf{Size}} & \multicolumn{2}{c}{\textbf{General}} & \multicolumn{2}{c}{\textbf{Math}} & \multicolumn{2}{c}{\textbf{Code}} & \multicolumn{2}{c}{\textbf{Reasoning}} & \multirow{2}{*}{\textbf{Average}} \\ 
        \cmidrule(r){3-4} \cmidrule(r){5-6} \cmidrule(r){7-8} \cmidrule(r){9-10}
         & & MMLU & IFEval & GSM8K & MATH & HumanEval & MBPP & Hellaswag & GPQA & \\
        \midrule
        Alpaca & 52.0$k$ & 47.46 & 41.09 & 35.63 & 4.96 & 39.63 & 37.40 & 48.11 & 5.56 & 32.48 \\
        \noalign{\vskip 0.5ex}\cdashline{1-11}\noalign{\vskip 0.5ex}
        \multicolumn{11}{c}{\textit{Data Selection}} \\
        \noalign{\vskip 0.5ex}
        Alpaca-clean & 51.7$k$ & 47.21 & 43.92 & 43.90 & 4.20 & 29.27 & 43.40 & 60.17 & 5.56 & 34.70 \\
        Superfiltering & 7.8$k$ & 39.96 & 37.80 & 44.50 & 5.38 & 40.85 & 44.00 & 42.38 & 27.27 & 35.27 \\
        Superfiltering GPT4 & 7.8$k$ & 37.71 & 34.35 & 53.68 & 11.00 & 9.15 & 45.60 & 57.81 & 2.53 & 31.48 \\
        Long & 1.0$k$ & 25.51 & 14.75 & 56.33 & 16.56 & 13.41 & 45.60 & 25.83 & 0.00* & 24.75 \\
        AlpaGasus & 9.2$k$ & 33.98 & 48.82 & 43.82 & 6.06 & 35.98 & 42.40 & 44.50 & 18.18 & 34.22 \\
        \noalign{\vskip 0.5ex}\cdashline{1-11}\noalign{\vskip 0.5ex}
        \multicolumn{11}{c}{\textit{Data Augmentation}} \\
        \noalign{\vskip 0.5ex}
        I-SHEEP & 8.4$k$ & 23.61 & 29.61 & 43.14 & 8.28 & 32.32 & 32.60 & 41.83 & 0.00* & 26.42 \\
        Alpaca-GPT4 & 52.0$k$ & 51.94 & 38.68 & 50.87 & 10.28 & 17.07 & 43.60 & 63.02 & 0.51 & 34.50 \\ 
        WizardLM & 70.0$k$ & 46.12 & 46.14 & 53.30 & 12.72 & 40.24 & 48.00 & 53.05 & 12.12 & 38.96 \\
        \midrule
         \textbf{\textit{Middo Alpaca}} & 57.6$k$ & 51.32 & 43.20 & 51.18 & 12.92 & 39.63 & 41.80 & 58.78 & 16.67 & 39.63 \\
         \textbf{\textit{MiddOnly$^\dagger$ Alpaca}} & 8.8$k$ & 43.47 & 40.78 & 65.20 & 15.58 & 51.83 & 47.60 & 58.65 & 17.68 & 42.60 \\
         \textbf{\textit{Middo Alpaca-4o-mini}} & 63.1$k$ & 44.69 & 47.96 & 57.62 & 18.50 & 52.44 & 45.40 & 57.37 & 19.70 & \textbf{42.96} \\
         \textbf{\textit{MiddOnly$^\dagger$ Alpaca-4o-mini}} & 24.9$k$ & 41.50 & 45.66 & 60.80 & 20.06 & 46.34 & 48.00 & 55.01 & 24.75 & \underline{42.77} \\
        \bottomrule
        \end{tabular}
    }
    \caption{Results of \method~compared to other baseline methods. The best and second best results are highlighted in \textbf{bold} and \underline{underlined}, respectively. Our method outperforms all baselines in the average score. *Note that 0.00 indicates that the method did not solve any examples. $\dagger$ Denotes training solely on Middo-generated data.}
    \label{tab:compare}
\end{table*}

We compare \method with both data selection (Alpaca-clean\citep{githubGitHubGururiseAlpacaDataCleaned}, Superfiltering~\citep{li-etal-2024-superfiltering}, Long~\citep{10.5555/3692070.3694581}, AlpaGasus~\citep{chen2024alpagasus}) and data augmentation (Alpaca-GPT4~\citep{peng2023instruction}, I-SHEEP~\citep{anonymous2025isheep}, WizardLM~\citep{xu2024wizardlm}) methods on the Alpaca dataset.

We use the optimal dataset obtained through \method from Alpaca for comparison with other baselines.
Additionally, to ensure a relatively fair comparison with data selection methods, we include a dataset that only uses the optimized data without incorporating any unoptimized samples, referred to as \textit{MiddOnly}, to isolate the effect of the optimization process and make a direct comparison with data selection approaches.

Results in Table~\ref{tab:compare} show our method achieves the highest average score of $42.96$, outperforming all other approaches.
Notably, even when using only the optimized subset \textit{MiddOnly Alpaca}, our method delivers a robust average score of $42.60$.
This demonstrates that iterative improvement is not primarily driven by data size, but rather by the effectiveness of our dynamic data selection and optimization process in identifying and generating data with high learning value for models.

\section{Analysis}

\subsection{Ablation Studies}

To assess the effectiveness of \method and the contribution of each optimization pipeline, we conduct ablation experiments with the LLaMA-3.1-8B model on the Alpaca dataset.
Specifically, we analyze the following ablations:
\textbf{(a) w/o loss}: removes \textit{Loss Patterns}.
\textbf{(b) w/o neighbor}: excludes \textit{Embedding Cluster Dynamics}.
\textbf{(c) w/o score}: removes \textit{Self-alignment Scores}.

\begin{wraptable}{r}{0.5\textwidth}
    \centering
    \caption{Ablation study on the development set. We report the performance of the model with different ablations. The ablations include removing the \textit{loss patterns}, \textit{embedding cluster dynamics} and \textit{self-alignment scores} separately. The best performance is highlighted in \textbf{bold}.}
    \label{tab:ablation}
    \resizebox{\linewidth}{!}{%
    \begin{tabular}{c l c c c c c}
        \toprule
        \textbf{Iter.} & \textbf{Ablations} & \textbf{IFEval} & \textbf{MATH} & \textbf{HumanEval} & \textbf{Hellaswag} & \textbf{Avg.} \\
        \midrule
        \multirow{4}{*}{\textit{iter1}} 
            & w             & 45.77 & 10.62 & 40.24 & 56.37 & \textbf{38.25} \\
            & w/o loss      & 42.49 & 10.11 & 39.02 & 59.53 & 37.79 \\
            & w/o neighbor  & 39.01 & 10.82 & 42.07 & 57.86 & 37.45 \\
            & w/o score     & 43.48 & 10.20 & 36.59 & 48.40 & 34.67 \\
        \midrule
        \multirow{4}{*}{\textit{iter2}} 
            & w             & 44.63 & 12.40 & 39.63 & 59.22 & \textbf{38.97} \\
            & w/o loss      & 42.28 & 9.92  & 42.68 & 58.21 & 38.27 \\
            & w/o neighbor  & 46.75 & 10.26 & 34.76 & 46.66 & 34.61 \\
            & w/o score     & 44.18 & 11.76 & 39.02 & 51.38 & 36.58 \\
        \midrule
        \multirow{4}{*}{\textit{iter3}} 
            & w             & 44.24 & 12.92 & 39.63 & 59.25 & \textbf{39.01} \\
            & w/o loss      & 43.18 & 12.42 & 36.59 & 55.30 & 36.87 \\
            & w/o neighbor  & 40.12 & 12.46 & 34.15 & 56.83 & 35.89 \\
            & w/o score     & 45.17 & 7.92  & 40.85 & 54.67 & 37.15 \\
        \bottomrule
    \end{tabular}%
    }
    \vspace{-20pt}
\end{wraptable}

The ablation results in Table~\ref{tab:ablation} consistently show that removing any part of the framework leads to a decline in performance across multiple iterations, reinforcing that each component plays a significant role in the overall performance.
This trend holds across the second (\textit{iter2}) and third (\textit{iter3}) iterations, where the removal of any pipeline consistently results in suboptimal performance, further highlighting the importance of balancing complexity, diversity, and quality in the optimization process.
These findings underscore the necessity of the full framework for achieving optimal results.

\subsection{Effect of Selected Data Scale}

\begin{wrapfigure}{l}{0.5\textwidth}
  \centering
  \includegraphics[width=\linewidth]{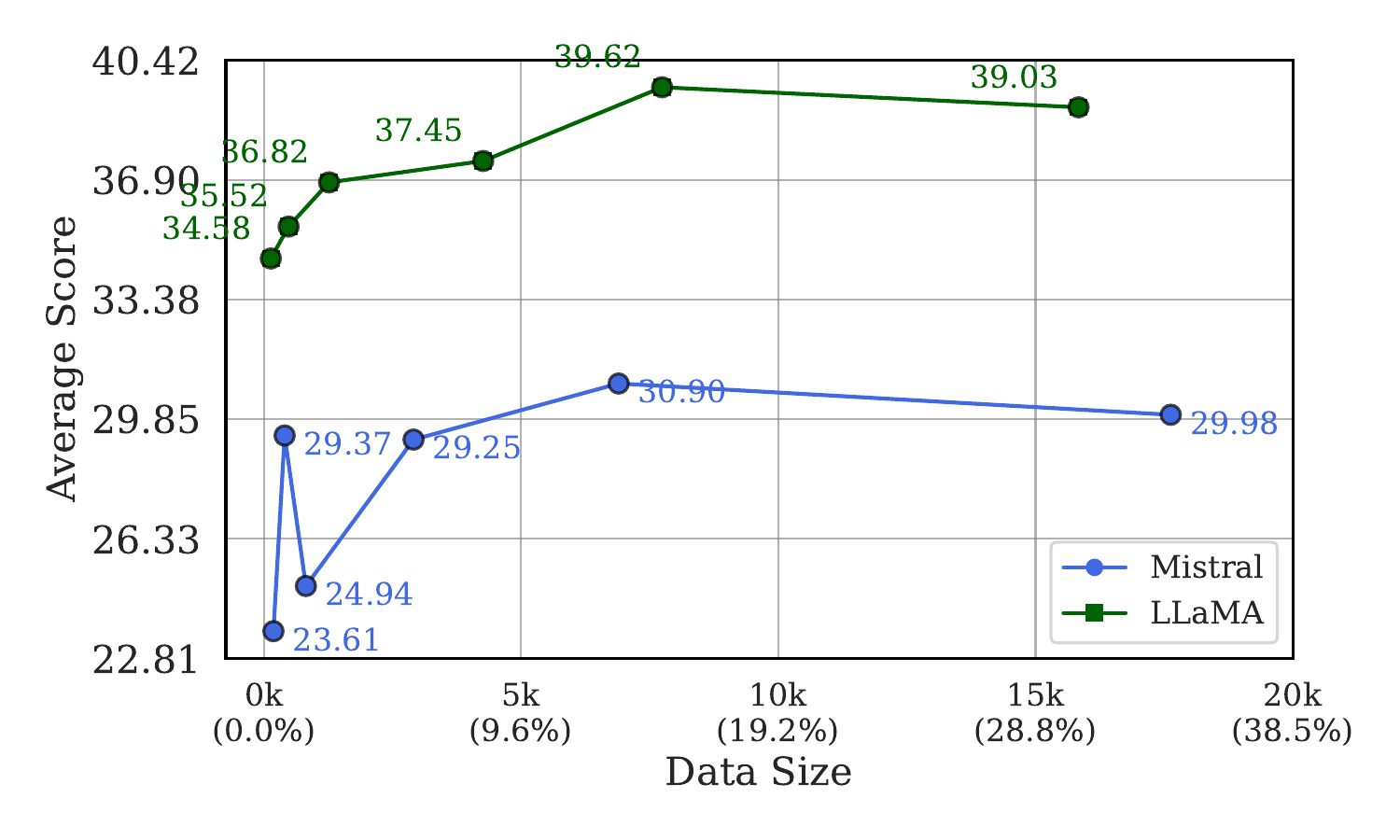}
  \caption{Performance comparison of \method on the Alpaca dataset with varying refined data sizes. The x-axis represents the number and percentage of data selected for refinement, while the y-axis shows the average accuracy across three iterations. To ensure fairness, we guarantee that the data after refinement is the same.}
  \label{fig:data-scale}
\end{wrapfigure}

We investigate the impact of the different scales of the selected and optimized data in this section by varying the thresholds for data selection.
Results are illustrated in Figure~\ref{fig:data-scale}.
We observe that increasing the size of the refined data initially leads to an upward trend in performance; however, once the refined data exceeds a certain threshold, performance begins to decline.
To maintain the potential for further iterative improvement, we set the refined data size at a moderate level that optimally balances the cost and benefit of the optimization process.
In the first iteration, each component selects approximately $5\%$ of the data for refinement.
By controlling the parameter $m$, the amount of data refined can adaptively change as the model’s capability increases.
Detailed data sizes selected in each iteration are provided in Appendix~\ref{sec:case}.

\subsection{Data Analysis}

\label{sec:effect_of_selected_data_scale}

\paragraph{Dynamic Iterative Improvement.}

For a deeper understanding of how \method transforms the dataset, we provide an analysis of its impact on data complexity, diversity, and quality.

\begin{figure}[htbp]
  \centering
  \begin{minipage}[t]{0.3\linewidth}
    \centering
    \includegraphics[width=\linewidth]{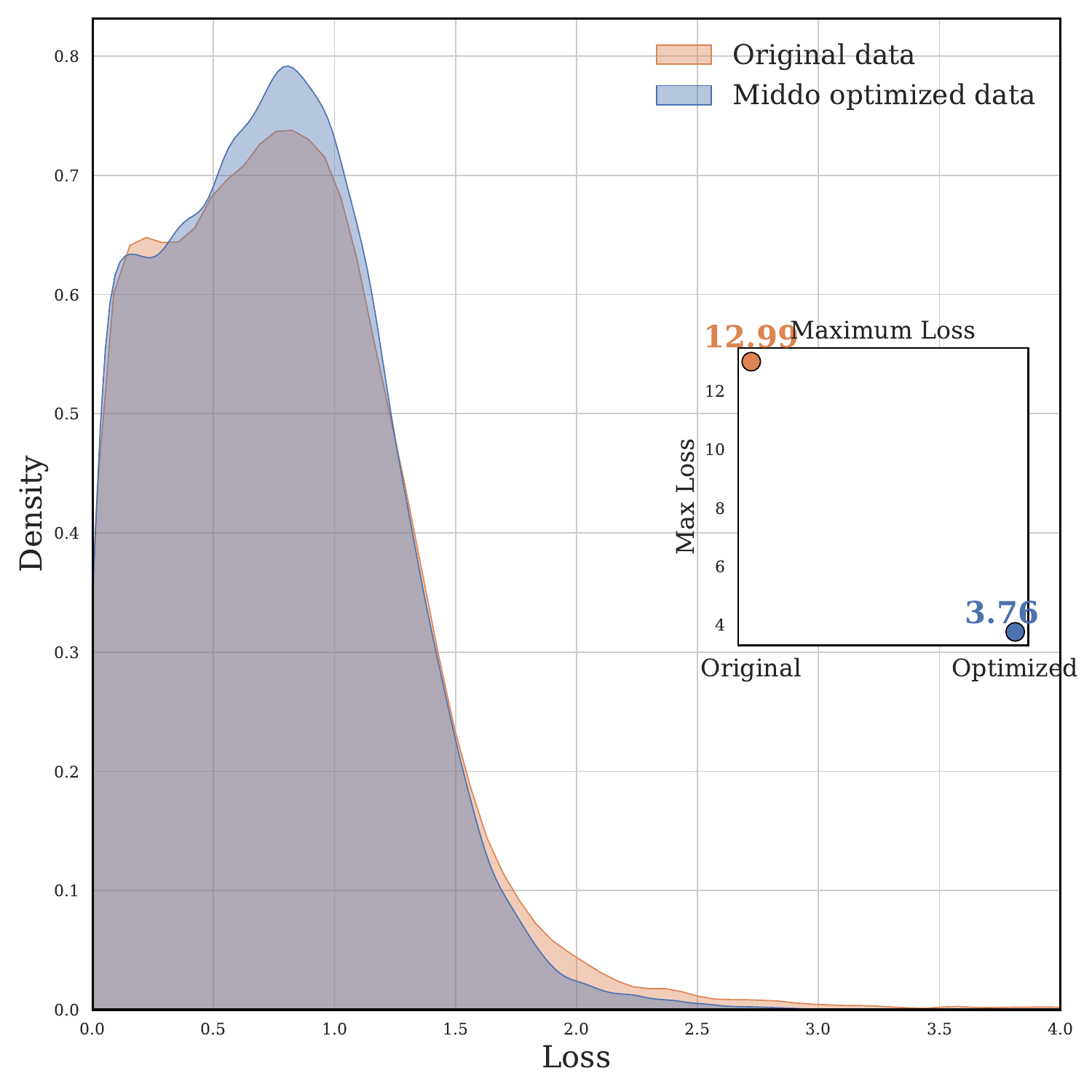}
    \captionof{figure}{Loss distribution comparison before and after applying \method. The density curve reflects the relative frequency of data points within specific loss intervals. The inset subfigure highlights the maximum loss reduction from $12.99$ to $3.76$.}
    \label{fig:loss-distribute}
  \end{minipage}
  \hfill
  \begin{minipage}[t]{0.315\linewidth}
    \centering
    \includegraphics[width=\linewidth]{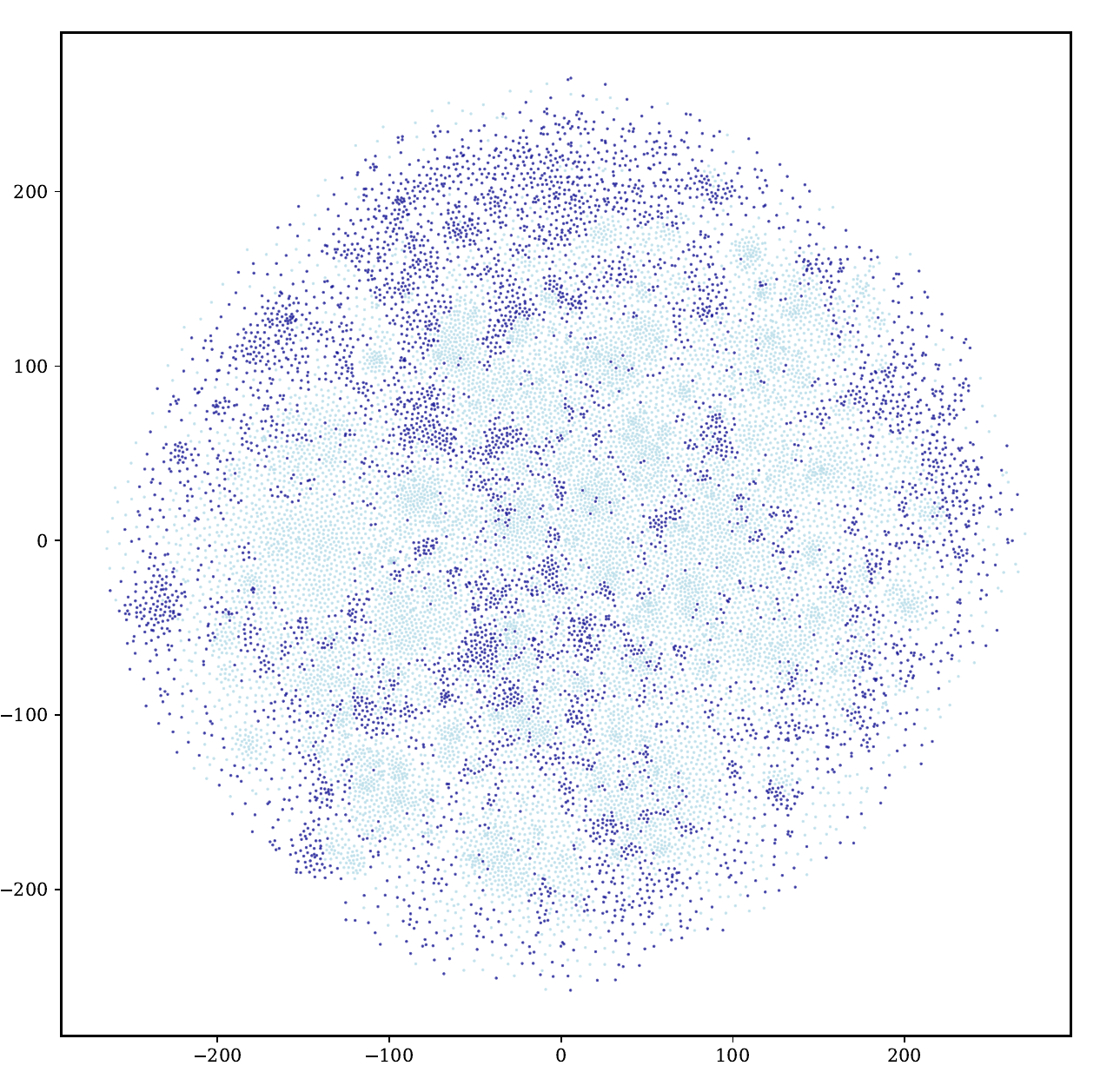}
    \captionof{figure}{t-SNE visualization of the Alpaca dataset before and after applying \method. The original dataset is shown in light blue, while the augmented data is in dark blue. The dark blue points tend to occupy the sparsely populated regions of the light blue point distribution.}
    \label{fig:tsne}
  \end{minipage}
  \hfill
  \begin{minipage}[t]{0.32\linewidth}
    \centering
    \includegraphics[width=\linewidth]{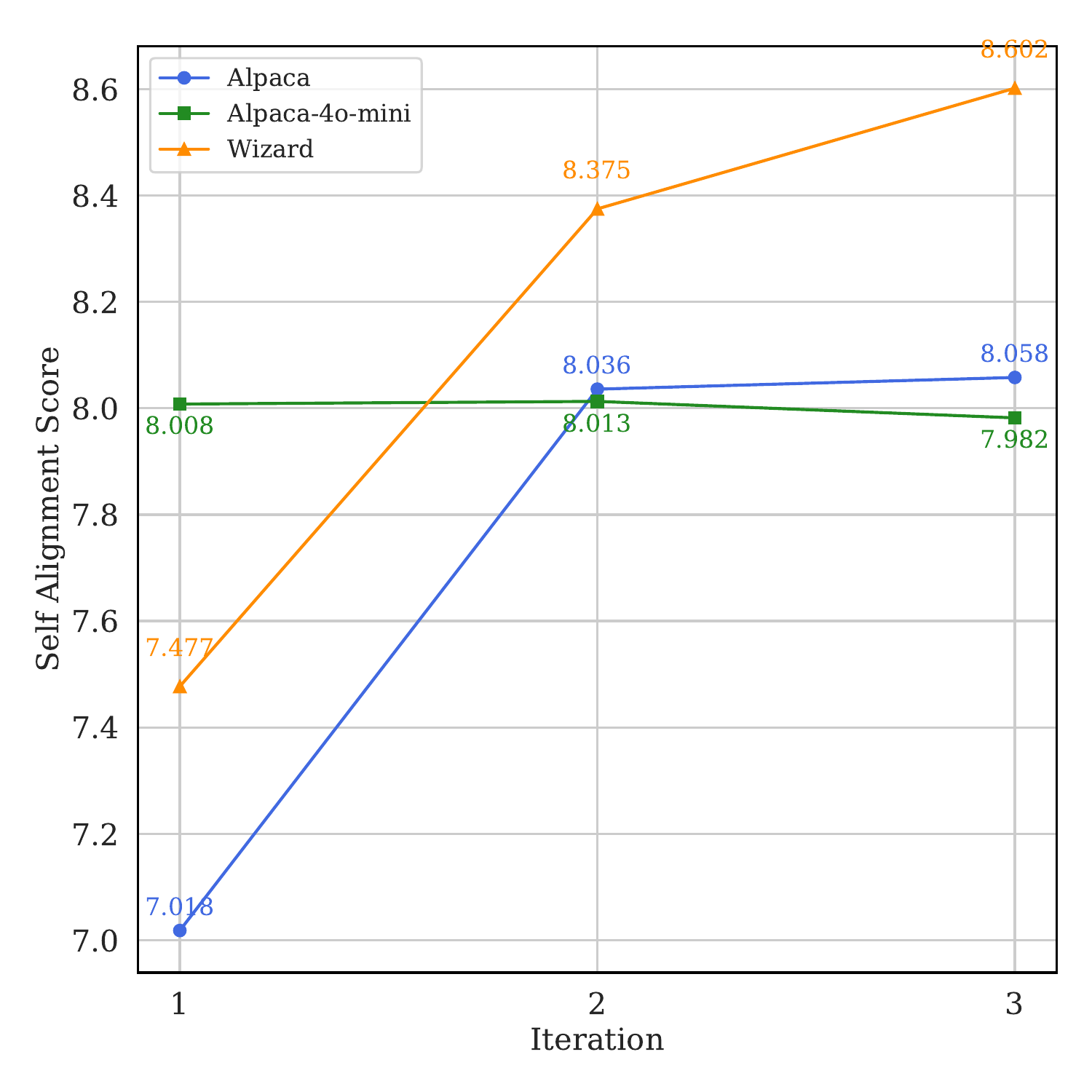}
    \captionof{figure}{Self-alignment score evolution across iterations. The x-axis represents the number of iterations, while the y-axis shows the average self-alignment score.}
    \label{fig:score}
  \end{minipage}
\end{figure}

\paragraph{Complexity.}
To quantify how \method modulates dataset complexity, we analyze the loss distribution evolution through optimization cycles.
As shown in Figure~\ref{fig:loss-distribute}, the original dataset exhibits a long-tailed distribution with extreme loss values up to $12.99$.
After applying \method, the maximum loss decreases by $71.05\%$ to $3.76$, indicating successful mitigation of overly complex samples and the distribution mode shifts leftward, suggesting better alignment between data complexity and model capability.
This transformation demonstrates our framework’s ability to adaptively prune pathological samples while preserving pedagogically valuable challenges.

\paragraph{Diversity.}

To analyze the diversity of the dataset after applying \method, we visualize the data distribution using t-SNE~\citep{tsne}.
Figure~\ref{fig:tsne} reveals how the augmented data points are distributed relative to the original data.
Notably, most of the augmented samples are located at the peripheries of the clusters, effectively filling in the sparsely populated regions.
This distribution indicates that \method is not merely adding redundant data but is instead enhancing the overall coverage of the latent space.
By strategically augmenting the dataset at the cluster edges, \method improves the diversity and ensures a more uniform distribution of data points, ultimately contributing to better model generalization.

\paragraph{Quality.}

The self-alignment score trajectories across different iterations are presented in Figure~\ref{fig:score}.
The observed trend indicates a gradual increase in the average score as the iterations progress.
This improvement signifies that the quality of the data is becoming more closely aligned with the model’s evolving capabilities.
Through the adversarial self-play mechanisms and iterative quality refinement, the model is able to assess and enhance the quality of both the instructions and responses within the dataset.
As the self-alignment scores increase, it reflects that the refined data is not only more accurate but also more consistent with the model’s internal standards and expectations.
This detailed evolution of the self-alignment scores provides critical insights into the dynamic process of dataset optimization, confirming that our approach effectively transforms low-quality samples into high-quality learning material over successive iterations. 

\section{Conclusion}

In this paper, we present \method, a model-informed dynamic data optimization framework that transforms LLM fine-tuning via closed-loop learning.
Unlike traditional static methods, \method establishes a self-evolving system that continuously adapts to the model’s evolving capabilities.
It employs three core mechanisms: complexity optimization refines overly complex samples using \textit{loss patterns}, ensuring the training data remains appropriately challenging; diversity optimization enhances dataset diversity by analyzing \textit{embedding cluster dynamics}; and quality optimization leverages \textit{self-alignment scores} to evaluate and improve the quality of training samples.
Experiments on multiple benchmarks demonstrate that \method  consistently boosts LLMs’ performance, achieving an average accuracy improvement of $7.15\%$ while maintaining the original data scale on LLaMA-3.1-8B. Ablation studies confirm the effectiveness of each component, underscoring the importance of balancing complexity, diversity, and quality.
\method’s adaptability and model-awareness make it a powerful tool for sustainable LLM training. Moreover, our approach paves the way for future research in adaptive training that continuously optimizes learning efficiency.

\section*{Limitations}

Despite its promising results, \method has several limitations:
(1) \method relies on the model being fine-tuned itself for identifying data quality and complexity. This means that the approach requires a sufficiently capable base model, and the performance may be limited if the base model is not strong enough to generate meaningful diagnostics for data refinement.
(2) \method does not currently utilize Reinforcement Learning, which could further enhance data refinement, especially for complex or subjective tasks.
(3) The closed-loop optimization system may lead to higher computational costs as the dataset grows or updates become more frequent, presenting scalability challenges.
(4) \method may propagate biases present in the initial training data, limiting fairness and generalization if the base model is trained on biased data.
These limitations highlight areas for future improvement, such as integrating RL, optimizing for scalability, and addressing data biases.

\section*{Acknowledgments}

This work is supported by Shanghai Artificial Intelligence Laboratory. Zinan Tang is an intern at Shanghai Artificial Intelligence Laboratory.

\bibliographystyle{plainnat}
\setcitestyle{numbers}
\bibliography{paper}

\clearpage
\newpage
\beginappendix

\section{Computational Cost Analysis}

\label{sec:computational_cost}

We analyzed the computational cost of Middo's optimization stages on 7B parameter models (LLaMA-3.1-8B, Mistral-7B-v0.3) using 50k-100k sample datasets (Alpaca, Alpaca-4o-mini, WizardLM) on 8 $\times$ NVIDIA A100 GPUs.

Each optimization iteration, encompassing data selection via \textit{loss patterns}, \textit{embedding cluster dynamics}, and \textit{self-alignment scores}, followed by refinement, typically completes in under 30 minutes. This efficiency is largely due to the parallelizable nature of the diagnostic modules and the use of acceleration techniques: CUDA for neighbor computation in \textit{Embedding Cluster Dynamics} and vLLM~\citep{10.1145/3600006.3613165} for batched inference during \textit{Self-alignment Score} calculation. Table~\ref{tab:overhead_cost} provides a detailed time breakdown per component, underscoring Middo's practical efficiency.

\begin{table*}[htbp]
\centering
\resizebox{\textwidth}{!}{%
\begin{tabular}{lcc}
\toprule
\textbf{Method Component} & \textbf{Time (Single A100 GPU)} & \textbf{Time (8 $\times$ A100 GPUs, Data Parallelism)} \\
\midrule
Loss patterns & $\approx$ 50 minutes & $\approx$ 10 minutes \\
Embedding cluster dynamics & $\approx$ 40 minutes + neighbor computation time & $\approx$ 10 minutes (CUDA acceleration) \\
Self-alignment scores & $\approx$ 1 hour per metric (6 metrics) & $\approx$ 10 minutes (vLLM acceleration) \\
\bottomrule
\end{tabular}
}
\caption{Approximate overhead time cost of \method's optimization components per iteration. Timings are reported for processing datasets in the range of 50k-100k samples.}
\label{tab:overhead_cost}
\end{table*}

\section{Hyperparameters Analysis}
\label{sec:hyper_para_analy}

\subsection{The Impact of Neighbor Number}

\label{sec:neighbour}

\begin{table*}[!htb]
    \centering
    \resizebox{0.8\textwidth}{!}{
        \begin{tabular}{ccccccccc}
        \toprule
        $\bm{k}$ & \textbf{IFEval} & \textbf{GSM8K} & \textbf{MATH} & \textbf{HumanEval} & \textbf{MBPP} & \textbf{Hellaswag} & \textbf{ARC-c} & \textbf{Average} \\
        \midrule
        $1$  & 43.59 & 38.74 & 9.20 & 35.98 & 39.8 & 48.59 & 17.17 & 33.3 \\
        $2$  & 51.56 & 43.21 & 10.72 & 40.85 & 41.00 & 57.47 & 12.12 & \textbf{35.72} \\
        $3$  & 40.82 & 40.49 & 9.50 & 32.32 & 39.20 & 59.72& 8.59 & 32.95 \\
        \bottomrule
        \end{tabular}
    }
    \caption{Impact of the number of neighbors ($k$) in the Embedding Cluster Dynamics on \method~performance. The table shows the performance across various benchmarks for different values of $k$, indicating that $k=2$ yields the best overall average score.}
    \label{tab:neighbor}
\end{table*}

We also explore how the number of neighbors $k$ used in the \textit{Embedding Cluster Dynamics} affects the overall performance of \method. By varying the number of neighbors, we analyze its impact on dataset diversity and model performance. Table~\ref{tab:neighbor}presents the results of this analysis. We find that the optimal number of neighbors is $k=2$, which achieves the best balance between diversity and performance. This setting ensures that the dataset is sufficiently expanded to enhance model generalization while avoiding excessive noise that may degrade performance.

\subsection{The Impact of Iterations}

\label{sec:iter}

\begin{wrapfigure}{r}{0.5\textwidth}
  \vspace{-20pt}
  \centering
  \includegraphics[width=\linewidth]{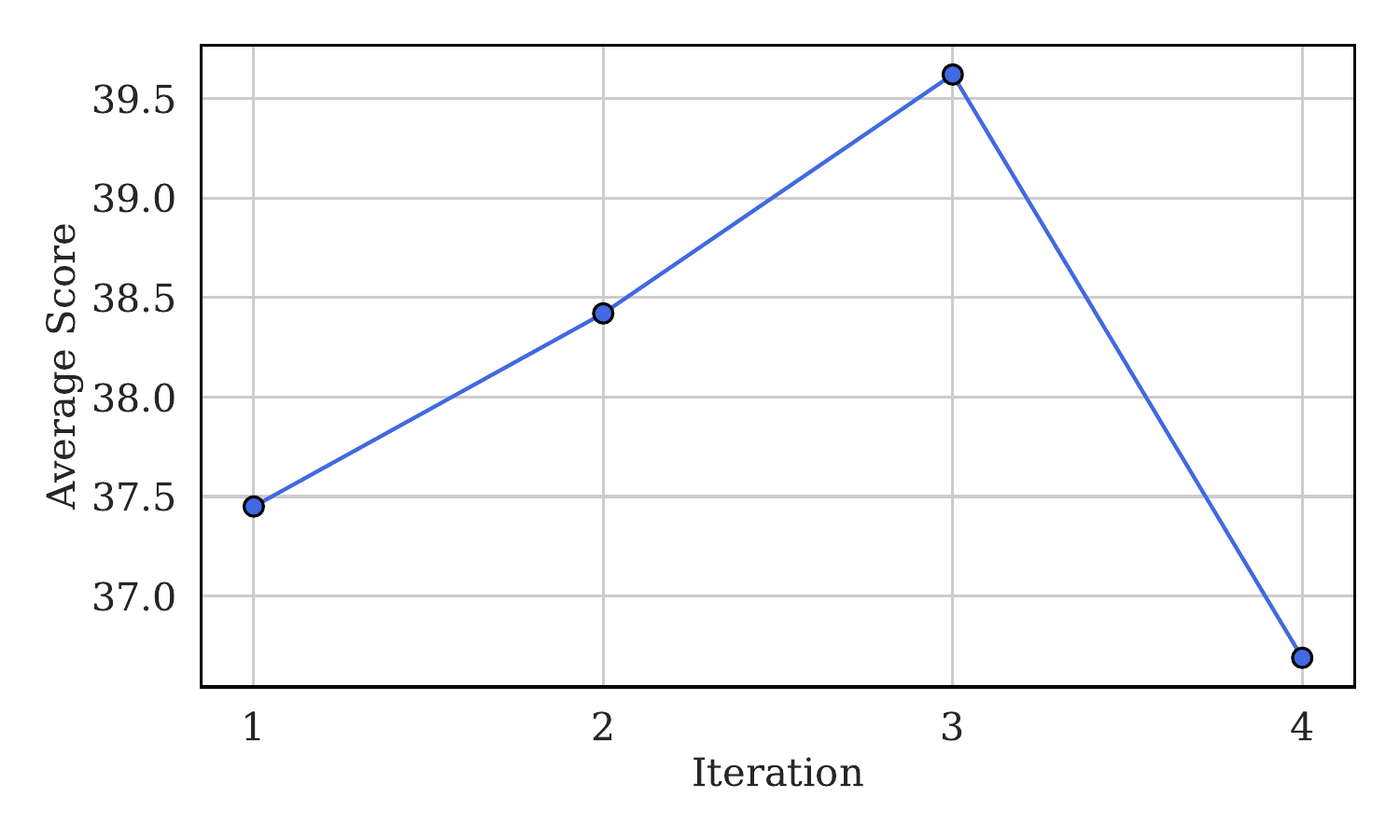}
  \caption{Performance trends on the Alpaca dataset across different iterations. The model’s performance peaks at three iterations.}
  \label{fig:iter}
  \vspace{-10pt}
\end{wrapfigure}

As shown in Figure~\ref{fig:iter}, we tested the number of iterations on the Alpaca dataset and found that the model's performance significantly declined after the third iteration. Therefore, we chose to optimize each dataset for three iterations. This optimal number is not necessarily fixed and may vary depending on the threshold of each iteration.

\subsection{The Impact of Thresholds}
\label{apx:thr}

The amount of data selected for refinement by each module (Loss Patterns \textit{(Complexity)}, Embedding Cluster Dynamics \textit{(Diversity)}, and Self-alignment Scores \textit{(Quality)}) is governed by dynamic thresholds $\tau = \mu + m\sigma$, where $\mu$ and $\sigma$ are the mean and standard deviation of the respective signal values (loss, diversity score, quality score) across the dataset. The multiplier $m$ is a key hyperparameter that controls the stringency of these thresholds.

Our approach to setting $m$ is guided by empirical analysis aimed at optimizing refinement effectiveness. Initial experiments (detailed in Section~\ref{sec:effect_of_selected_data_scale}, Figure~\ref{fig:data-scale}) indicated that refining a total unique proportion of approximately $10-20\%$ of the dataset in the first iteration yields substantial performance improvements.

To determine appropriate $m$ values for each module, we conducted a sensitivity analysis, presented in Table~\ref{tab:m_sensitivity}. This table shows how different combinations of $m$ for complexity, diversity, and quality impact the total percentage of unique data selected for refinement and the resulting average model performance on benchmarks. The $m$ values are varied (e.g., in $0.5$ increments) for each signal, and combinations are chosen to target the 10-20\% total selected data range. As shown, performance peaks when the selected data proportion falls within this empirically determined optimal range. For instance, the combination yielding $14.88\%$ selected data achieved the best average score of $43.23$. When multiple $m$ combinations meet the $10-20\%$ criterion, we select those with the smallest absolute $m$ values (representing the mildest effective thresholds) that achieve this target, balancing refinement impact with efficiency.

\begin{table}[htbp]
    \centering
    \begin{tabular}{c c c c c c c}
    \toprule
    \textbf{Dataset} & \textbf{iteration} & \textbf{\textit{loss}} & \textbf{\textit{neighbor}} & \textbf{\textit{self}} & \textbf{total} \\
    \midrule
    \multicolumn{6}{c}{\textit{LLaMA-3.1-8B}} \\
        \noalign{\vskip 0.5ex}
    \multirow{4}{*}{\textbf{Alpaca}} & \textit{init} & $m=1$ & $m=-1$ & $m=-1.5$ & 52,002 \\
    & \textit{iter1} & 1180 & 1924 & 1159 & 53,939 \\
    & \textit{iter2} & 299 & 1853 & 108 & 55,811 \\
    & \textit{iter3} & 242 & 1822 & 381 & 57,636 \\
    \midrule
    \multirow{4}{*}{\shortstack{\textbf{Alpaca} \\ \textbf{4o-mini}}} & \textit{init} & $m=0$ & $m=-1$ & $m=-0.5$ & 52,002 \\
    & \textit{iter1} & 5684 & 8032 & 4145 & 60,865 \\
    & \textit{iter2} & 611 & 2291 & 876 & 63,184 \\
    & \textit{iter3} & 472 & 2127 & 661 & 65,324 \\
    \midrule
    \multirow{4}{*}{\textbf{Wizard}} & \textit{init} & $m=1$ & $m=-1.5$ & $m=-2$ & 70,000 \\
    & \textit{iter1} & 3585 & 3585 & 2690 & 73,642 \\
    & \textit{iter2} & 959 & 3341 & 1016 & 76,993 \\
    & \textit{iter3} & 751 & 3414 & 420 & 80,419 \\
    \specialrule{1.5pt}{0pt}{0pt}
    \multicolumn{6}{c}{\textit{Mistral-7B-v0.3}} \\
        \noalign{\vskip 0.5ex}
    \multirow{4}{*}{\textbf{Alpaca}} & \textit{init} & $m=0.5$ & $m=-2$ & $m=-1$ & 52,002 \\
    & \textit{iter1} & 2418 & 2111 & 2367 & 54,131 \\
    & \textit{iter2} & 1985 & 2091 & 932 & 56,268 \\
    & \textit{iter3} & 1788 & 1982 & 352 & 58,348 \\
    \midrule
    \multirow{4}{*}{\shortstack{\textbf{Alpaca} \\ \textbf{4o-mini}}} & \textit{init} & $m=1$ & $m=-2$ & $m=-2.5$ & 52,002 \\
    & \textit{iter1} & 1407 & 7691 & 1499 & 59,696 \\
    & \textit{iter2} & 1278 & 9116 & 1045 & 68,874 \\
    & \textit{iter3} & 1346 & 2487 & 661 & 74,036 \\
    \midrule
    \multirow{4}{*}{\textbf{Wizard}} & \textit{init} & $m=1$ & $m=-1.5$ & $m=-1.5$ & 70,000 \\
    & \textit{iter1} & 5637 & 5709 & 5258 & 76,429 \\
    & \textit{iter2} & 3558 & 5999 & 6310 & 82,501 \\
    & \textit{iter3} & 3885 & 6229 & 3767 & 89,178 \\
    \bottomrule
    \end{tabular}
    \caption{Data Size Details Across Iterative Refinement. For each dataset, the table lists the number of samples selected by the three components—\textit{loss} (\textit{Loss Patterns}), \textit{neighbor} (\textit{Embedding Cluster Dynamics}), and \textit{self} (\textit{Self-alignment Scores}). During each iteration, along with the total data size after refinement. The \textit{init} row represents the original dataset size and the threshold controlling hyperparameter $m$ corresponding to each component.}
    \label{tab:size}
\end{table}

\begin{table}[htbp]
\centering
\begin{tabular}{cccccc}
\toprule
\textbf{Complexity} & \textbf{Diversity} & \textbf{Quality} & \textbf{Total Selected} & \textbf{Percentage} & \textbf{Performance} \\
\midrule
$m=0$ & $m=-1$ & $m=-1.5$ & 15.8k & 30.45\% & 41.81 \\
$m=0.5$ & $m=-1.5$ & $m=-1.5$ & 7.7k & 14.88\% & \textbf{43.23} \\
$m=1$ & $m=-2$ & $m=-1.5$ & 4.3k & 8.20\% & 41.96 \\
$m=1.5$ & $m=-2.5$ & $m=-1.5$ & 2.6k & 5.09\% & 41.55 \\
$m=2$ & $m=-3$ & $m=-4$ & 1.3k & 2.44\% & 40.87 \\
$m=3$ & $m=-3.5$ & $m=-10$ & 0.5k & 0.92\% & 39.64 \\
$m=4$ & $m=-4$ & $m=-12$ & 0.1k & 0.26\% & 38.69 \\
\bottomrule
\end{tabular}
\caption{Sensitivity analysis for the threshold multiplier $m$ on the Alpaca dataset (first iteration). The table shows the impact of varying $m$ for complexity, diversity, and quality modules on the total unique data selected (sum and percentage) and the average model performance (mean score across benchmarks).}
\label{tab:m_sensitivity}
\vspace{-0.3cm}
\end{table}

The actual data sizes selected in each iteration for the experiments reported in the main paper, using $m$ values derived from this sensitivity analysis (e.g., targeting the $15\%$ mark initially), are detailed in Table~\ref{tab:size}. As the model's performance improves over subsequent iterations, the amount of data flagged by these fixed $m$ thresholds naturally decreases due to shifts in the signal distributions ($\mu$ and $\sigma$). This adaptive selection aligns with our observation that early training phases benefit from addressing a broader set of initial complexities and diversities, while later stages refine more nuanced aspects.

We do not place excessive emphasis on the improvements brought about by differences in data volume, so our selection may not necessarily be optimal.

\section{Experimental Details}

\subsection{Instruction Fine-tune Dataset}

We evaluate \method on three general instruction fine-tuning datasets.
\begin{itemize}
    \item \textbf{Alpaca}~\citep{alpaca}: consists of 52,002 instruction-response pairs generated by Stanford University using the self-instruct~\citep{wang-etal-2023-self-instruct} method based on OpenAI's text-davinci-003. This dataset is designed for fine-tuning dialogue models similar to ChatGPT to achieve efficient instruction-following capabilities.
    \item \textbf{Alpaca-4o-mini}: to evaluate performance on a higher-quality response dataset, we generated responses for all Alpaca instructions using GPT-4o mini, creating the Alpaca-4o-mini dataset.
    \item  \textbf{WizardLM}~\citep{xu2024wizardlm}: 70K data generated based on Evol-Instruct, which aims to generate more complex instruction data through a recursive evolutionary approach in order to improve the model's reasoning and instruction comprehension.
\end{itemize}

\subsection{Models}

We primarily conducted experiments on LLaMA 3.1-8B, and additionally performed extra experiments on Mistral 7B-v0.3.
\begin{itemize}
  \item \textbf{LLaMA 3.1-8B}~\citep{grattafiori2024llama3herdmodels}: LLaMA 3.1-8B is a large language model released by Meta, featuring 8 billion (8B) parameters. It is part of the LLaMA (Large Language Model Meta AI) series, focusing on efficient reasoning and text generation capabilities. LLaMA 3.1-8B excels in code generation, language understanding, and conversational tasks, optimizing inference speed and training efficiency, making it suitable for research, commercial applications, and AI studies.
  \item \textbf{Mistral 7B-v0.3}~\citep{jiang2023mistral7b}: Mistral 7B-v0.3 is an open-source language model developed by Mistral AI, featuring 7 billion parameters. It is optimized based on the Transformer architecture, emphasizing efficiency and multitasking capabilities. Compared to earlier versions, this model shows improvements in coding, mathematics, and reasoning tasks, making it suitable for chatbots, programming assistance, and natural language processing tasks. Mistral 7B-v0.3 incorporates feedback from the open-source community to enhance inference efficiency, delivering high performance with reduced computational resources.
\end{itemize}

\subsection{Benchmarks}

We assess model performance on general knowledge, mathematical problem-solving, code generation and commonsense reasoning benchmarks.
\begin{itemize}
  \item \textbf{IFEval (Instruction Following Evaluation)}~\citep{zhou2023instructionfollowingevaluationlargelanguage}: a benchmark dataset designed to assess the instruction-following capabilities of large models. It encompasses various tasks, including general knowledge question answering, commonsense reasoning, and mathematical reasoning, aiming to measure the understanding and accuracy of language models when executing complex instructions.
  \item \textbf{MMLU (Massive Multitask Language Understanding)}~\citep{hendrycks2021measuring}: a large-scale, multi-task language understanding benchmark that covers 57 subjects, testing models on their knowledge and reasoning abilities across fields such as history, law, mathematics, and medicine. It serves as a significant indicator of general artificial intelligence knowledge levels.
  \item \textbf{GSM8K (Grade School Math 8K)}~\citep{cobbe2021trainingverifierssolvemath}: a dataset specifically created for solving mathematical problems, containing approximately 8,500 elementary school math questions that primarily focus on basic arithmetic, logical reasoning, and text comprehension skills. This dataset is used to evaluate models' mathematical computation and reasoning abilities.
  \item \textbf{MATH}~\citep{hendrycks2021measuring2}: consists of math competition problems from middle school and college levels, covering areas such as algebra, geometry, number theory, and calculus. This dataset is more challenging than GSM8K and is primarily used to assess models' performance on advanced mathematical reasoning tasks.
  \item \textbf{HumanEval}~\citep{chen2021evaluatinglargelanguagemodels}: a dataset for evaluating code generation capabilities, featuring a series of Python programming problems, each with a clear function signature and test cases. This dataset is commonly used to measure AI performance in automated code generation and programming tasks.
  \item \textbf{MBPP (Mostly Basic Programming Problems)}~\citep{austin2021programsynthesislargelanguage}: a benchmark dataset for code generation, containing 1,000 basic programming questions that cover data structures, algorithms, and logical reasoning. It is suitable for assessing AI capabilities in fundamental programming tasks.
  \item \textbf{Hellaswag}~\citep{zellers-etal-2019-hellaswag}: a benchmark dataset for commonsense reasoning, consisting of a series of incomplete sentences that require models to select the most reasonable ending. This dataset tests models' contextual understanding and reasoning abilities by designing misleading options.
  \item \textbf{GPQA (Graduate-Level Google-Proof Q\&A)}~\citep{rein2024gpqa}: a challenging dataset designed to evaluate the capabilities of LLMs and scalable oversight mechanisms. Let me provide more details about it.
\end{itemize}

\subsection{Baselines} 

We compare \method with both existing data selection and data augmentation methods on the Alpaca dataset.

\paragraph{Data Selection Methods.}
\begin{itemize}
  \item \textbf{Alpaca-clean}~\citep{githubGitHubGururiseAlpacaDataCleaned}: a cleaned version of the Alpaca dataset that removes low-quality samples and duplicates, aiming to improve the overall quality of the dataset.
  \item \textbf{Superfiltering}~\citep{li-etal-2024-superfiltering}: using smaller, weaker language models (such as GPT-2) as data filters to compute IFD allows for the selection of high-quality instruction tuning data.
  \item \textbf{Long}~\citep{10.5555/3692070.3694581}: directly select the 1,000 samples with the longest responses as training data.
  \item \textbf{AlpaGasus}~\citep{chen2024alpagasus}: utilize powerful LLMs (such as ChatGPT) to automatically assess the sample quality in the Alpaca dataset and filter out high-quality data to enhance model training effectiveness.
\end{itemize}

\paragraph{Data Augmentation Methods.}
\begin{itemize}
  \item \textbf{Alpaca-GPT4}~\citep{peng2023instruction}: a data augmentation method that uses GPT-4 to generate additional training data for the Alpaca dataset.
  \item \textbf{I-SHEEP}~\citep{anonymous2025isheep}: a data augmentation method that uses a self-supervised learning approach to generate additional training data for the Alpaca dataset.
  \item \textbf{WizardLM}~\citep{xu2024wizardlm}: 70K data generated based on Evol-Instruct, which aims to generate more complex instruction data through a recursive evolutionary approach in order to improve the model's reasoning and instruction comprehension.
\end{itemize}

\subsection{Hyperparameters}
\label{sec:hyper}

\paragraph{Fine-tune.} For LLaMA-3.1-8B, we follow the Alpaca GitHub repository\footnote{\url{https://github.com/tatsu-lab/stanford_alpaca}}, setting the batch size to 32, the learning rate to $2 \times 10^{-5}$, and the warmup ratio to 0.03. For Mistral-7B-v0.3, we adjust the learning rate to $1 \times 10^{-5}$, as per official recommendations\footnote{\url{https://docs.mistral.ai/capabilities/finetuning}}. All the hyperparameters are detailed in Table~\ref{tab:hyper_llama} and Table~\ref{tab:hyper_mistral}. 

\paragraph{Data Synthetic.} We use the OpenAI API to generate data by GPT-4o-mini, setting both temperature and top\_p to 1.0 to guarantee diversity.

\begin{table}[htbp]
    \centering
    \begin{minipage}{0.365\textwidth}
        \centering
        \resizebox{\textwidth}{!}{\begin{tabular}{lc}
            \toprule
            \textbf{Hyperparameter} & \textbf{Value} \\
            \midrule
            \multicolumn{2}{c}{\textit{LLaMA-3.1-8B}} \\
            Learning Rate & $2 \times 10^{-5}$ \\
            Number of Epochs & $1$ \\
            Number of Devices & $8$ \\
            Per-device Batch Size & $4$ \\
            Gradient Accumulation Steps & $8$ \\
            Learning Rate Scheduler & cosine \\
            Warmup Ratio & $0.03$ \\
            Max Sequence Length & $4096$ \\
            \bottomrule
        \end{tabular}}
        \caption{LLaMA-3.1-8B SFT Hyperparameters.}
        \label{tab:hyper_llama}
    \end{minipage}
    \begin{minipage}{0.365\textwidth}
        \centering
        \resizebox{\textwidth}{!}{\begin{tabular}{lc}
            \toprule
            \textbf{Hyperparameter} & \textbf{Value} \\
            \midrule
            \multicolumn{2}{c}{\textit{Mistral-7B-v0.3}} \\
            Learning Rate & $1 \times 10^{-5}$ \\
            Number of Epochs & $1$ \\
            Number of Devices & $8$ \\
            Per-device Batch Size & $4$ \\
            Gradient Accumulation Steps & $8$ \\
            Learning Rate Scheduler & cosine \\
            Warmup Ratio & $0.03$ \\
            Max Sequence Length & $4096$ \\
            \bottomrule
        \end{tabular}}
        \caption{Mistral-7B-v0.3 SFT Hyperparameters.}
        \label{tab:hyper_mistral}
    \end{minipage}
    \begin{minipage}{0.23\textwidth}
        \centering
        \resizebox{\textwidth}{!}{\begin{tabular}{lc}
            \toprule
            \textbf{Hyperparameter} & \textbf{Value} \\
            \midrule
            Pass@n & $n=1$ \\
            Presence Penalty & $0.0$ \\
            Frequency Penalty & $0.0$ \\
            Repetition Penalty & $1.0$ \\
            Temperature & $0.0$ \\
            Top\_p & $1.0$ \\
            Top\_k & $-1$ \\
            Min\_p & $0.0$ \\
            Max Tokens & $4096$ \\
            Min Tokens & $0$ \\
            \bottomrule
        \end{tabular}}
        \caption{Evaluation Hyperparameters.}
        \label{tab:eval_hyper}
    \end{minipage}
\end{table}

\paragraph{Evaluation.} All benchmarks are conducted in zero-shot and we conducted the tests using the default configuration of OpenCompass. All the hyperparameters are detailed in Table~\ref{tab:eval_hyper}.

All experiments are conducted on 8 $\times$ NVIDIA Tesla A100 GPUs about 50 GPU hours.

\section{Self-alignment Scores}

\begin{figure}[htbp]
  \centering
  \begin{subfigure}[b]{0.3\textwidth}
    \centering
    \includegraphics[width=\linewidth]{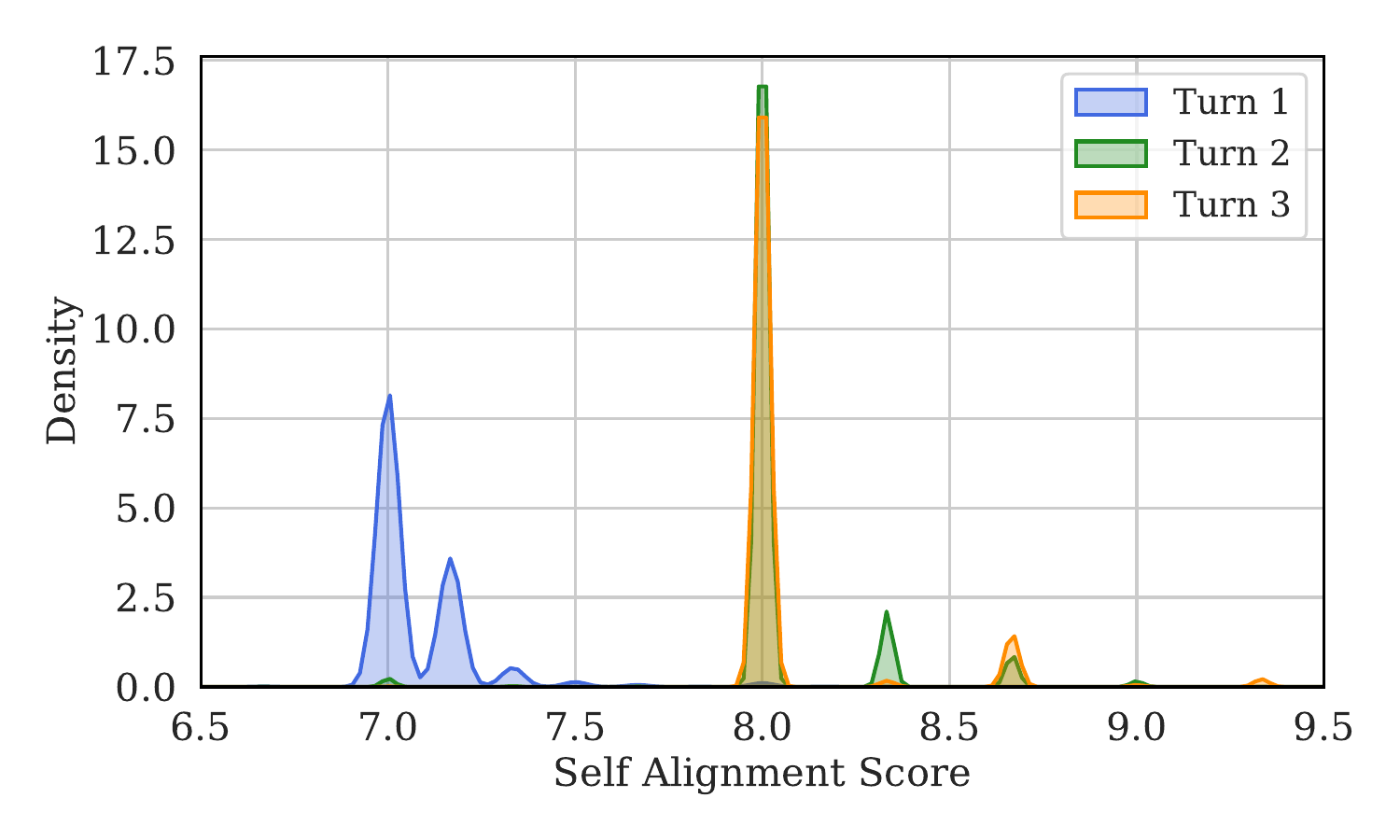}
    \caption{Alpaca dataset.}
    \label{fig:alpaca_score}
  \end{subfigure}
  \hfill
  \begin{subfigure}[b]{0.3\textwidth}
    \centering
    \includegraphics[width=\linewidth]{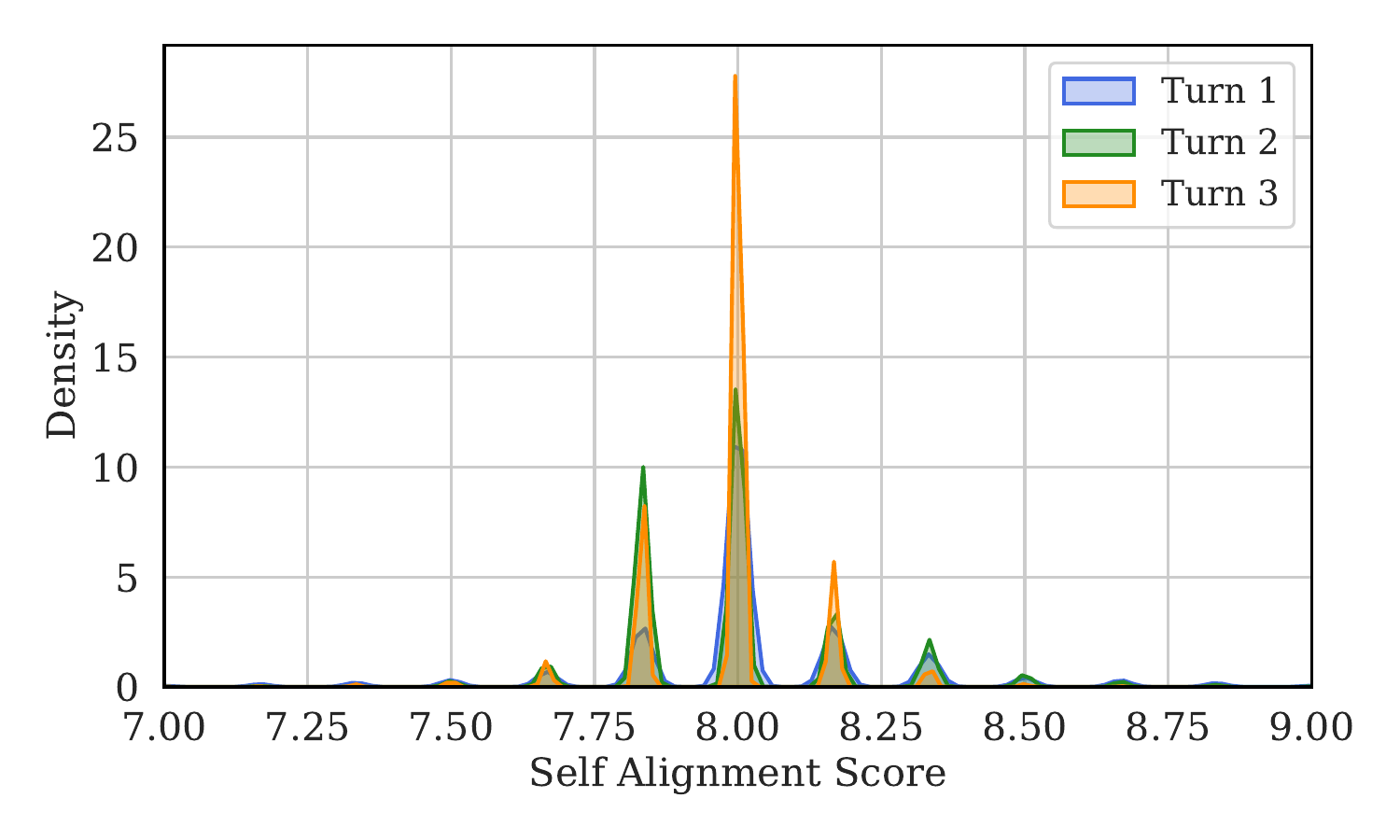}
    \caption{Alpaca-4o-mini dataset.}
    \label{fig:alpaca_4o_mini_score}
  \end{subfigure}
  \hfill
  \begin{subfigure}[b]{0.3\textwidth}
    \centering
    \includegraphics[width=\linewidth]{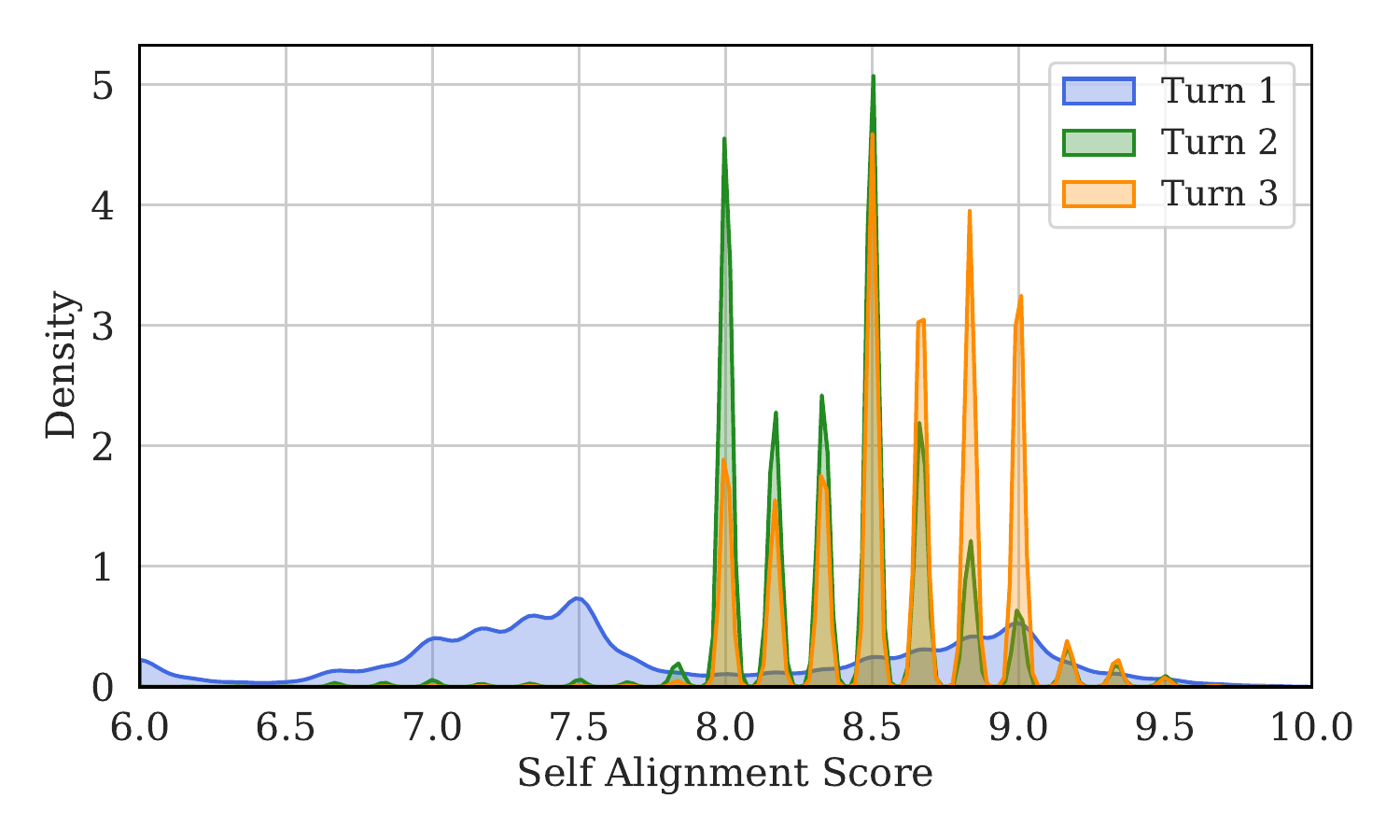}
    \caption{WizardLM dataset.}
    \label{fig:wizardlm_score}
  \end{subfigure}
  \caption{Self-alignment score evolution across iterations. The x-axis represents the self-alignment scores, while the y-axis shows the density of data points.}
  \label{fig:self_score}
\end{figure}

We provide detailed self-alignment score evolution across iterations on the Alpaca, Alpaca-4o-mini, and WizardLM datasets in Figure~\ref{fig:self_score}. These figures illustrate the dynamic evolution of self-alignment scores across iterations, highlighting the continuous improvement in dataset quality and alignment with model capabilities.

\section{Case Study}

\label{sec:case}

\begin{figure}[!htb]
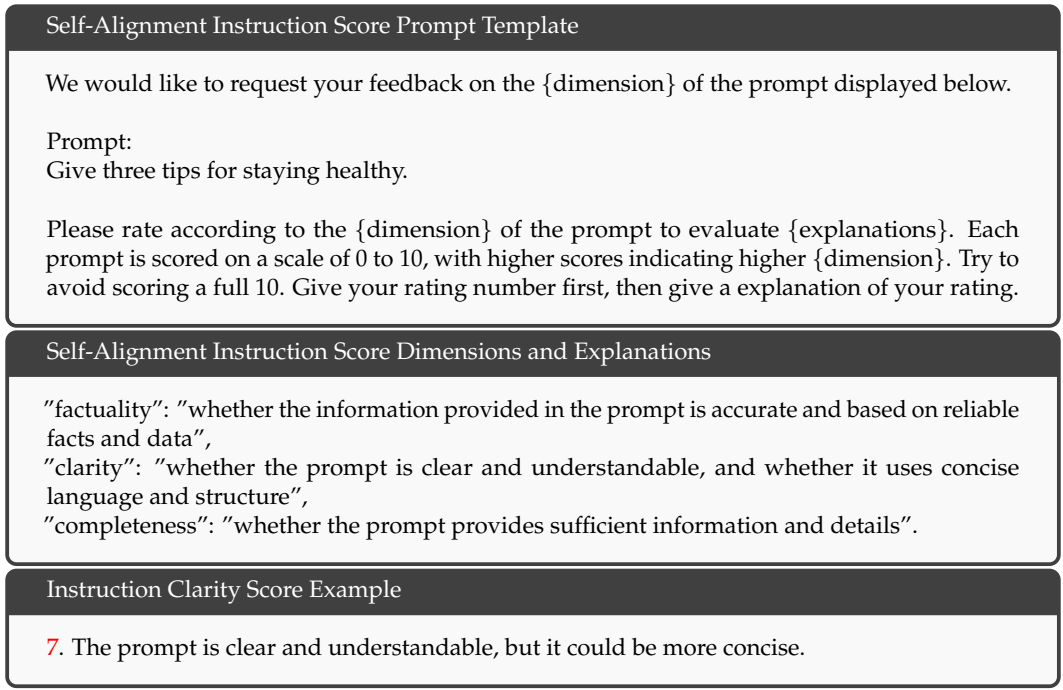

    \centering
    \small
    \begin{minipage}{0.9\textwidth}
      \begin{tcolorbox}[colback=gray!5,colframe=black!75, width=\textwidth, title=Self-Alignment Instruction Score Prompt Template]
        We would like to request your feedback on the \{dimension\} of the prompt displayed below.\\

        Prompt:\\
        Give three tips for staying healthy.\\

        Please rate according to the \{dimension\} of the prompt to evaluate \{explanations\}. Each prompt is scored on a scale of 0 to 10, with higher scores indicating higher \{dimension\}. Try to avoid scoring a full 10. Give your rating number first, then give a explanation of your rating.
        \end{tcolorbox}
    \end{minipage}
    \\
    \begin{minipage}{0.9\textwidth}
      \begin{tcolorbox}[colback=gray!5,colframe=black!75, width=\textwidth, title=Self-Alignment Instruction Score Dimensions and Explanations]
    "factuality": "whether the information provided in the prompt is accurate and based on reliable facts and data",\\
    "clarity": "whether the prompt is clear and understandable, and whether it uses concise language and structure",\\
    "completeness": "whether the prompt provides sufficient information and details".
        \end{tcolorbox}
    \end{minipage}
    \\
    \begin{minipage}{0.9\textwidth}
      \begin{tcolorbox}[colback=gray!5,colframe=black!75, width=\textwidth, title=Instruction Clarity Score Example]
    {\color{red}7}. The prompt is clear and understandable, but it could be more concise.
        \end{tcolorbox}
    \end{minipage}
    \caption{Self-Alignment instruction score example.}
  \end{figure}

  \begin{figure}[!htb]
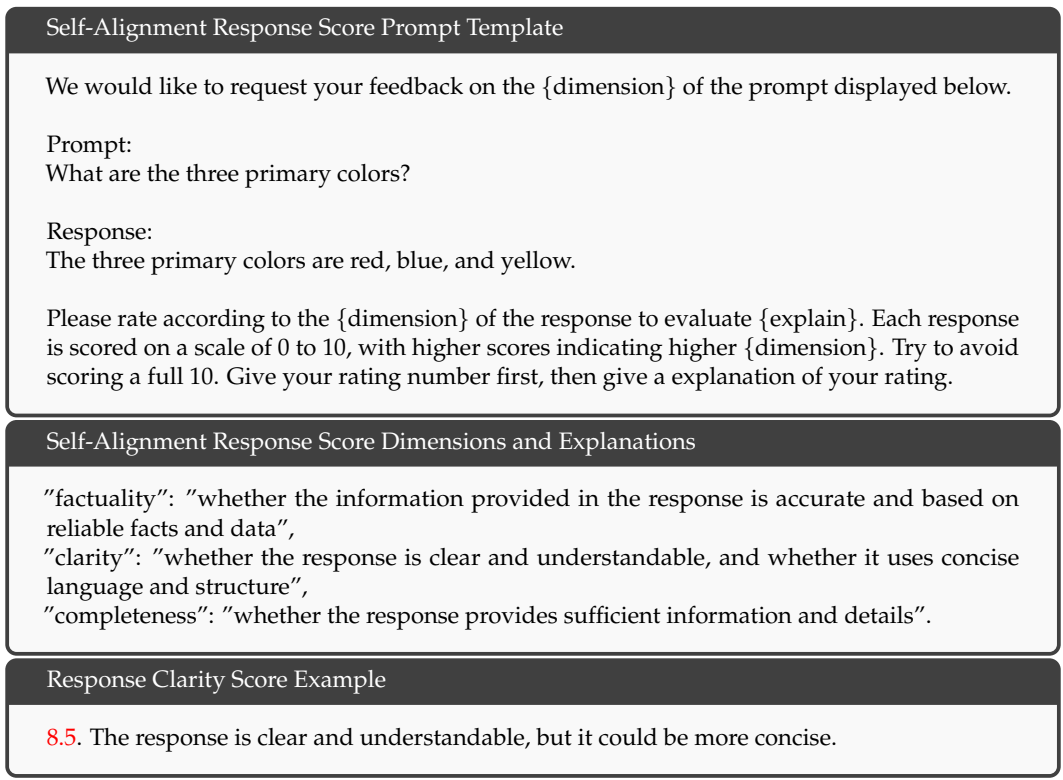

    \centering
    \small
    \begin{minipage}{0.9\textwidth}
      \begin{tcolorbox}[colback=gray!5,colframe=black!75, width=\textwidth, title=Self-Alignment Response Score Prompt Template]
        We would like to request your feedback on the \{dimension\} of the prompt displayed below.\\

        Prompt:\\
        What are the three primary colors?\\

        Response:\\
        The three primary colors are red, blue, and yellow.\\

        Please rate according to the \{dimension\} of the response to evaluate \{explain\}. Each response is scored on a scale of 0 to 10, with higher scores indicating higher \{dimension\}. Try to avoid scoring a full 10. Give your rating number first, then give a explanation of your rating.
        \end{tcolorbox}
    \end{minipage}
    \\
    \begin{minipage}{0.9\textwidth}
      \begin{tcolorbox}[colback=gray!5,colframe=black!75, width=\textwidth, title=Self-Alignment Response Score Dimensions and Explanations]
    "factuality": "whether the information provided in the response is accurate and based on reliable facts and data",\\
    "clarity": "whether the response is clear and understandable, and whether it uses concise language and structure",\\
    "completeness": "whether the response provides sufficient information and details".
        \end{tcolorbox}
    \end{minipage}
    \\
    \begin{minipage}{0.9\textwidth}
      \begin{tcolorbox}[colback=gray!5,colframe=black!75, width=\textwidth, title=Response Clarity Score Example]
    {\color{red}8.5}. The response is clear and understandable, but it could be more concise.
        \end{tcolorbox}
    \end{minipage}
    \caption{Self-Alignment response score example.}
  \end{figure}

\begin{figure}[htbp]
    \centering
    \small
    \begin{minipage}{0.9\textwidth}
      \begin{tcolorbox}[colback=gray!5,colframe=black!75, width=\textwidth, title=Complexity Optimization Prompt Template]
        You are a powerful LLM with the task to simplify the given \#Prompt\# for weaker open source LLMs (e.g. LLaMA, Qwen). Since the original \#Prompt\# is hard for them to handle, you need to rewrite it into a simpler version that these weaker LLMs can handle or learn from more easily. Follow the steps below carefully.

        Step 1: Read the \#Prompt\# in detail. Suggest possible methods to make this prompt easier for weaker LLMs to handle or learn from, then list these methods as your \#Methods List\#.
        
        Step 2: Create a comprehensive plan to simplify the \#Prompt\# using several methods from your \#Methods List\#. Present your plan in a clear, step-by-step format.
        
        Step 3: Execute your plan to rewrite the \#Prompt\# into a simpler, more learnable version. You can change scenarios, contexts, or settings as needed. Your goal is to ensure that weaker LLMs learn from this prompt, rather than just memorizing an answer.
        
        Step 4: Finally, review your rewritten version for any problems. Present only the \#Final Rewritten Prompt\#, without any additional explanation.
        
        Please respond strictly in the following format:
        
        Step 1 \#Methods List\#:
        
        Step 2 \#Plan\#:
        
        Step 3 \#Rewritten Prompt\#:
        
        Step 4 \#Final Rewritten Prompt\#:
        \end{tcolorbox}
    \end{minipage}
    \\
    \begin{minipage}{0.9\textwidth}
      \begin{tcolorbox}[colback=gray!5,colframe=black!75, width=\textwidth, title=Overly Hard Data Example]
        Find the mode of the following set of numbers: {\color{red}23, 16, 22, 19, 24, 21}
        \end{tcolorbox}
    \end{minipage}
    \\
    \begin{minipage}{0.9\textwidth}
      \begin{tcolorbox}[colback=gray!5,colframe=black!75, width=\textwidth, title=Optimize Steps]
        \textbf{Step 1 \#Methods List\#:}\\
        1. Simplify the language used in the prompt.\\
        2. Break down the instructions into simpler steps.\\
        3. Provide examples to clarify the concept of mode.\\
        4. Use smaller numbers or more relatable contexts.\\
        5. Ask the question in a more direct manner.\\
    
        \textbf{Step 2 \#Plan\#:}\\
        1. Change complex terminology like ``mode'' to a more descriptive explanation.\\
        2. Use a familiar context, like favorite fruits or animals.\\
        3. Clearly outline the steps needed to find the most common number, potentially using a smaller set.\\
        4. Offer a simple example to illustrate the mode calculation.\\
        5. Keep the structure straightforward and direct.\\
  
        \textbf{Step 3 \#Rewritten Prompt\#:}\\
        Find the most frequent number in this list: 3, 7, 2, 3, 5, 7. What number appears the most?'\\
  
        \textbf{Step 4 \#Final Rewritten Prompt\#:}\\
        Find the number that appears the most in this list: 3, 7, 2, 3, 5, 7. What is that number?
        \end{tcolorbox}
    \end{minipage}
    \\
    \begin{minipage}{0.9\textwidth}
      \begin{tcolorbox}[colback=gray!5,colframe=black!75, width=\textwidth, title=Optimized Easier Data]
        Find the number that {\color{red}appears the most} in this list: {\color{red}3, 7, 2, 3, 5, 7}. What is that number?
        \end{tcolorbox}
    \end{minipage}
    \caption{Complexity optimization example.}
    \label{fig:complexity}
  \end{figure}

\begin{figure}[htbp]
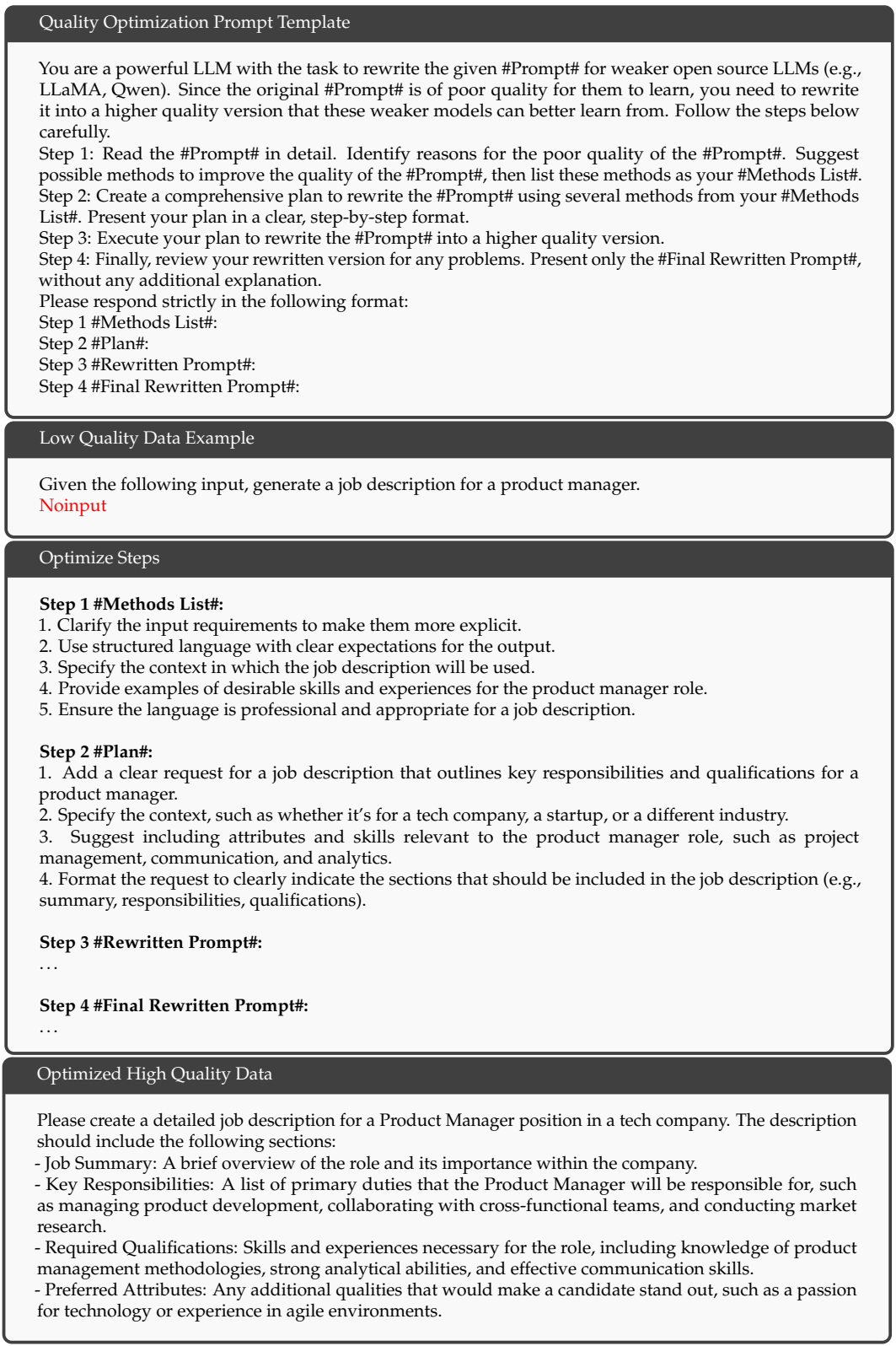

    \centering
    \footnotesize
    \begin{minipage}{0.9\textwidth}
      \begin{tcolorbox}[colback=gray!5,colframe=black!75, width=\textwidth, title=Quality Optimization Prompt Template]
        You are a powerful LLM with the task to rewrite the given \#Prompt\# for weaker open source LLMs (e.g., LLaMA, Qwen). Since the original \#Prompt\# is of poor quality for them to learn, you need to rewrite it into a higher quality version that these weaker models can better learn from. Follow the steps below carefully.
        
        Step 1: Read the \#Prompt\# in detail. Identify reasons for the poor quality of the \#Prompt\#. Suggest possible methods to improve the quality of the \#Prompt\#, then list these methods as your \#Methods List\#.
        
        Step 2: Create a comprehensive plan to rewrite the \#Prompt\# using several methods from your \#Methods List\#. Present your plan in a clear, step-by-step format.
        
        Step 3: Execute your plan to rewrite the \#Prompt\# into a higher quality version.
        
        Step 4: Finally, review your rewritten version for any problems. Present only the \#Final Rewritten Prompt\#, without any additional explanation.
        
        Please respond strictly in the following format:
        
        Step 1 \#Methods List\#:
        
        Step 2 \#Plan\#:
        
        Step 3 \#Rewritten Prompt\#:
        
        Step 4 \#Final Rewritten Prompt\#:
        \end{tcolorbox}
    \end{minipage}
    \\
    \begin{minipage}{0.9\textwidth}
      \begin{tcolorbox}[colback=gray!5,colframe=black!75, width=\textwidth, title=Low Quality Data Example]
        Given the following input, generate a job description for a product manager.\\
        {\color{red}{Noinput}}
        \end{tcolorbox}
    \end{minipage}
    \\
    \begin{minipage}{0.9\textwidth}
      \begin{tcolorbox}[colback=gray!5,colframe=black!75, width=\textwidth, title=Optimize Steps]
        \textbf{Step 1 \#Methods List\#:}\\
        1. Clarify the input requirements to make them more explicit.\\
        2. Use structured language with clear expectations for the output.\\
        3. Specify the context in which the job description will be used.\\
        4. Provide examples of desirable skills and experiences for the product manager role.\\
        5. Ensure the language is professional and appropriate for a job description.\\
        
        \textbf{Step 2 \#Plan\#:}\\
        1. Add a clear request for a job description that outlines key responsibilities and qualifications for a product manager.\\
        2. Specify the context, such as whether it's for a tech company, a startup, or a different industry.\\
        3. Suggest including attributes and skills relevant to the product manager role, such as project management, communication, and analytics.\\
        4. Format the request to clearly indicate the sections that should be included in the job description (e.g., summary, responsibilities, qualifications).\\
        
        \textbf{Step 3 \#Rewritten Prompt\#:}\\
        \dots \\
        
        \textbf{Step 4 \#Final Rewritten Prompt\#:}\\
        \dots
        \end{tcolorbox}
    \end{minipage}
    \\
    \begin{minipage}{0.9\textwidth}
      \begin{tcolorbox}[colback=gray!5,colframe=black!75, width=\textwidth, title=Optimized High Quality Data]
        Please create a detailed job description for a Product Manager position in a tech company. The description should include the following sections: \\
        - Job Summary: A brief overview of the role and its importance within the company.\\
        - Key Responsibilities: A list of primary duties that the Product Manager will be responsible for, such as managing product development, collaborating with cross-functional teams, and conducting market research.\\
        - Required Qualifications: Skills and experiences necessary for the role, including knowledge of product management methodologies, strong analytical abilities, and effective communication skills.\\
        - Preferred Attributes: Any additional qualities that would make a candidate stand out, such as a passion for technology or experience in agile environments.
        \end{tcolorbox}
    \end{minipage}
    \vspace{-1pt}
    \caption{Quality optimization example.}
    \label{fig:quality}
  \end{figure}

\begin{figure}[!t]
    \small
    \centering
    \begin{minipage}{0.9\textwidth}
      \begin{tcolorbox}[colback=gray!5,colframe=black!75, width=\textwidth, title=Diversity Extension Prompt Template]
        You are a powerful LLM with the task to create brand new prompts for weaker open source LLMs (e.g. LLaMA, Qwen). You need to create a brand new complete prompt for them to learn in order to improve their knowledge and skills. Follow the steps below carefully.\\
        Use \#Hint Prompt 1\# and \#Hint Prompt 2\# as guiding examples. Then read the \#Core Prompt\# in detail. Be inspired to suggest additional new prompts, and ultimately create only one completely original and diverse \#New Prompt\#.\\
        Please respond strictly in the following format:\\
        \#New Prompt\#:"
        \end{tcolorbox}
    \end{minipage}
    \\
    \begin{minipage}{0.9\textwidth}
      \begin{tcolorbox}[colback=gray!5,colframe=black!75, width=\textwidth, title=Sparse Data And Neighbors]
        \textbf{\#Hint Prompt 1\#:}\\
        How long did Shakespeare live?\\
        
        \textbf{\#Hint Prompt 2\#:}\\
        How did the Industrial Revolution change society?\\
        
        \textbf{\#Core Prompt\#:}\\
        How did Julius Caesar die?
        \end{tcolorbox}
    \end{minipage}
    \\
    \begin{minipage}{0.9\textwidth}
      \begin{tcolorbox}[colback=gray!5,colframe=black!75, width=\textwidth, title=Extensioned Data]
        What were the key factors that led to the fall of the Roman Empire?
        \end{tcolorbox}
    \end{minipage}
    \caption{Diversity extension example.}
    \label{fig:diversity}
  \end{figure}

\end{document}